\documentclass[lettersize, 10 pt, journal, twoside]{IEEEtran}
\usepackage{amsmath,amsfonts}
\usepackage{array}
\usepackage[font=normalsize,labelfont=sf,textfont=sf]{subcaption}
\usepackage{placeins}

\usepackage{textcomp}
\usepackage{url}
\usepackage{verbatim}
\usepackage{graphicx}
\usepackage{cite}
\usepackage{color,soul} 
\usepackage{float}
\usepackage{dblfloatfix}

\usepackage[ruled, longend, vlined, linesnumbered]{algorithm2e}
\usepackage{multirow}
\usepackage{enumitem}
\usepackage{booktabs}
\usepackage{makecell}
\usepackage{amssymb}
\usepackage{comment}
\setlist[description]{style=nextline}

\newcommand{\ie}{{\em i.e.,~}}

\IEEEaftertitletext{\vspace{-2.0\baselineskip}}

\begin{document}

\title{DevGRU: Depth-guided Visual Navigation using a Collision-aware Recurrent Model}

\author{Kyung~Min~Han,
        Eunsom~Kim,
        and~Young~J.~Kim%
\thanks{Accepted for publication in IEEE Robotics and Automation Letters
(RA-L), 2026. \copyright~2026 IEEE. Personal use of this material is
permitted. Permission from IEEE must be obtained for all other uses, in
any current or future media, including reprinting/republishing this
material for advertising or promotional purposes, creating new collective
works, for resale or redistribution to servers or lists, or reuse of any
copyrighted component of this work in other works.}%
}

\maketitle

\begin{abstract}
Existing {visual} navigation models often aim to develop foundation models that can generalize robot navigation across diverse platforms. However, many of these models are prone to collisions when deployed in complex indoor environments, particularly in structured layouts and narrow passages. To address this problem, we propose a depth {image- and point-goal-conditioned} navigation system, DevGRU. {The proposed system employs an action predictor (AP) that generates collision-aware future trajectories, enabling effective avoidance of immediate obstacles. In conjunction with a collision predictor, the AP further compensates for errors accumulated in the goal pose estimation and proactively mitigates future deviations.}  
{To evaluate our method, we conducted experiments across nine different scenes and three state-of-the-art approaches - ViNT, NoMaD, and NavDP - as well as four additional variants of ViNT and NoMaD}. In terms of navigation performance, DevGRU significantly outperforms ViNT and NoMaD by a large margin. In addition, the proposed model has a relatively small number of trainable parameters, resulting in the fastest inference time among the baselines, particularly outperforming NavDP by 7$\times$ in model size and 17$\times$ in inference time.
\end{abstract}

\begin{IEEEkeywords}
Vision-Based Navigation, Deep Learning Method, Collision Avoidance, Embodied AI
\end{IEEEkeywords}

\section{Introduction}

Traditional robot navigation relies on engineering pipelines composed of localization, metric mapping, and planning~\cite{HanKim20,MinKim23,HanKim22}, each of which has been studied for decades. While effective, such approaches often require accurate geometric maps and localization, which can be costly to build and maintain, especially in large, featureless, or evolving environments. This has motivated growing interest in learning-based visual navigation, which seeks to reduce reliance on dense metric maps by learning navigation behaviors directly from visual observations.

As an alternative to metric maps, topological map-based navigation has recently attracted increasing attention, where a topological map represents the environment as a graph of visual observations. This representation still provides compact long-range route guidance without explicit metric reconstruction. Recent state-of-the-art visual navigation methods, including GNM~\cite{ShaLev23gnm}, ViNT~\cite{ShaLev23vint}, and NoMaD~\cite{SriLev24}, adopt this topological navigation framework and combine it with learned visual policies for local planning. Furthermore, many of these studies aim to develop navigational foundation models that can generalize across diverse robot platforms with different sensor configurations and measurement modalities.

While we acknowledge the importance of developing general-purpose foundation models, we observe that their robustness is often insufficient for reliable deployment even in typical indoor environments. {These approaches rely solely on color images, which are unreliable for predicting obstacles, and they are predominantly trained on collision-free datasets, limiting their ability to handle collision-prone situations.}
Unlike outdoor spaces, which typically offer large, open areas for navigation, indoor environments often feature dense man-made structures and narrow corridors. 
As a result, robots may collide with obstacles before reaching the goal.  Such collisions can cause self-damage or lead the robot to become stuck, ultimately resulting in navigation failure. 
To address the limitations of existing approaches, we aim to develop an indoor navigation system with two key features: (1) collision-aware navigation and (2) a lightweight model trained with relatively small amount of dataset that enables fast inference. Both are crucial for mobile robot navigation, where safe operation depends on reliable obstacle avoidance and rapid action updates.

To improve collision awareness, we use depth images rather than color images, which are directly connected to environmental geometry and therefore provides more informative cues for obstacle avoidance. Specifically, we introduce a collision prediction model that estimates the likelihood of collision from the depth image. In addition, we propose an action-prediction model that predicts both the robot's trajectory and a {collision-aware goal} configuration, a key feature that distinguishes our system from existing models.

To support learning with a compact model under limited data, we collect a relatively small dataset comprising 10K collision-free and 10K collision navigation samples. Our navigation model is trained using a compact architecture based on the Gated Recurrent Unit (GRU)~\cite{ChoBen14}, enabling efficient learning with limited data. In contrast, {high-capacity models such as Transformers would suffer from overfitting in data-limited regimes~\cite{DosHou21}.} {Distillation or transfer learning from existing foundation models is also not applicable in our problem setting due to both modality (color vs depth) and task mismatch (collision-aware navigation). 
}

The proposed system, DevGRU, is evaluated in various indoor environments and compared with state-of-the-art baseline methods to assess navigation performance. In addition, we analyze the model size and measure inference time to evaluate the compactness and efficiency of our approach relative to the baseline methods.
Overall, the contributions of this paper are as follows:
\begin{itemize}
\item A collision-aware {indoor navigation framework that enables successful long-horizon traversal.}
\item A lightweight depth-based visual navigation model that can be trained with a relatively small dataset.
\item A new depth-image navigation dataset that includes collision-intensive navigation samples.
\item {Consistent performance gains over ViNT, NoMaD and their variants in terms of SPL (0.8 vs.\ 0.15), along with substantially reduced inference time (2.5$\times$, 3.5$\times$, and 17$\times$ faster than NoMaD, ViNT, and NavDP, respectively)}.
\end{itemize}

\section{Related Work}
\label{Related Work}

Embodied AI navigation has become an active research area in recent years, driven by advances in large-scale simulation platforms and learning-based navigation policies~\cite{SavBat19}\cite{ZhuFar17}. Early studies in this field were largely conducted in simulated environments and often relied on RL approaches, {with tasks defined by different types of goal specifications: Object-Goal, Image-Goal (IG), and Point-Goal (PG).} 
IG and PG navigation are particularly relevant to our work because they directly guide robots toward local target locations.
{Similar to existing visual navigation methods~\cite{ShaLev23vint,SriLev24,ShaLev21}, our approach follows an IG-conditioned navigation framework. However, unlike the previous approaches, we further incorporate PG navigation characteristics, as explored in~\cite{WanNej26,SavBat19}. This enables our system to explicitly estimate the geometric distance to the goal configuration, whereas pure IG methods rely solely on temporal distance to the goal.}


\subsection{Image-Goal Navigation}
IG navigation aims to drive an agent toward a target location implicitly specified by a goal image. In this setting, the navigation policy receives the goal image together with the current (and sometimes past) observations and predicts actions that move the agent toward the goal. Many existing studies adopt RGB images as the primary input modality. In simulated environments, IG navigation is often formulated as an RL problem in which the agent outputs discrete actions~\cite{SavBat19}\cite{MezBoj22}. {However, transferring a learned model to real-world robotic deployment is nontrivial due to a sim-to-real gap. 
}

By repeatedly executing IG navigation toward a series of intermediate image sub-goals (SGs), a robot can gradually reach the final goal. This strategy has been explored in several recent works~\cite{SriLev24}\cite{ShaLev23vint}\cite{SheMen25}. 
However, relying primarily on RGB images is often insufficient for navigating narrow indoor environments containing walls, doors, and pillars.

\subsection{Point-Goal Navigation}
PG navigation explicitly utilizes the target goal location. Mobile robots can estimate the target {PG} with respect to the current robot configuration using local odometry. 
Similar to the IG navigation setting, a sequence of sub-goals in terms of PG can be defined to implement a topological navigation strategy. A large number of PG-based navigation approaches have been proposed~\cite{HuaLi26}\cite{LiuYan24}\cite{RotHut24}. However, such approaches can be vulnerable during long-distance navigation in indoor environments due to accumulated odometry errors and uncertainties in the estimated poses. One possible solution is to improve the robot’s global localization accuracy~\cite{BonWol24}\cite{YanHut23}; however, doing so contradicts the original motivation of {mapless} navigation, which is to avoid relying on explicit localization. 
{Although our method also adopts the PG navigation paradigm, it alleviates pose error accumulation by predicting corrected PGs, enabling longer-horizon navigation.}

\subsection{Topological Navigation}
Topological navigation is a classical strategy in mobile robot navigation~\cite{SieSca11}. Unlike metric maps, which are based on geometric representations of the environment, topological maps provide a more compact, simpler representation that can scale efficiently to large environments. Due to their practical advantages, topological navigation strategies have been widely used in conventional mobile robot navigation and exploration~\cite{ChoNag01} and have recently been adopted in visual navigation~\cite {ShaLev23vint,SriLev24}, including our approach.



\section{The DevGRU model}
\label{DEVGRU model}

\subsection{ {Visual Navigation using a Topological Map} }
\label{sec:topo_nav}

\begin{figure}[b]
    \centering
    \includegraphics[width=0.95\linewidth]{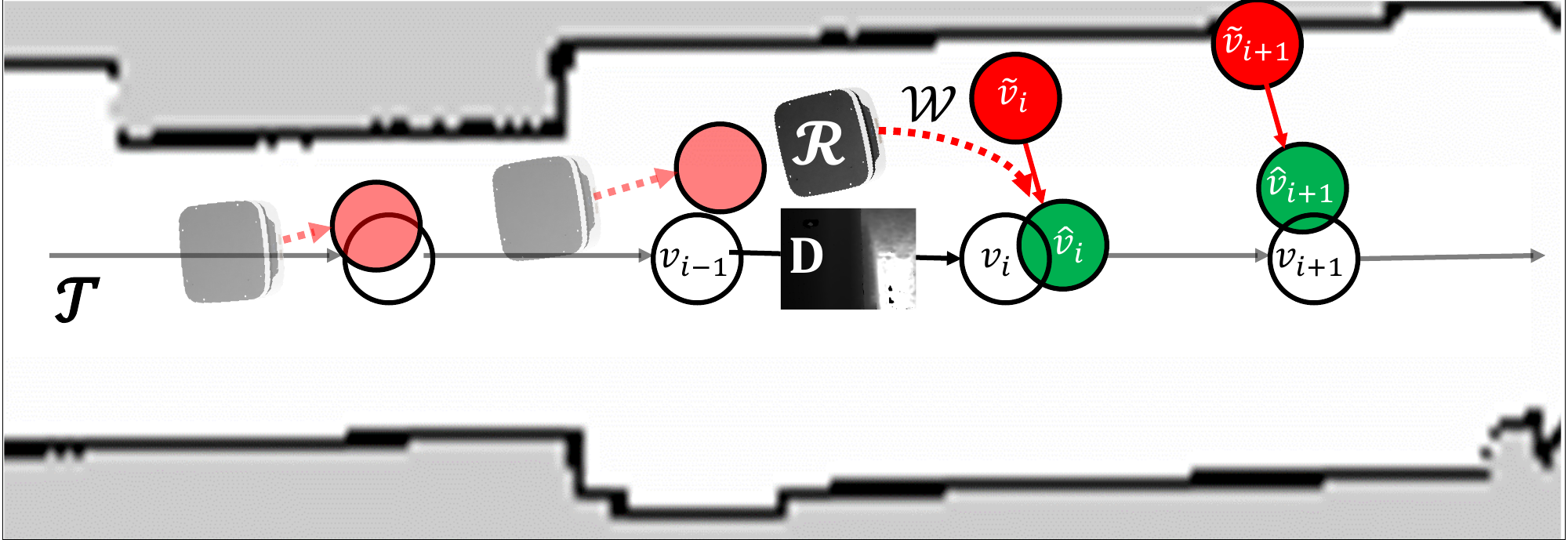}
    \captionsetup{skip=2pt}
    \caption{A mobile robot $\mathcal{R}$ moves from $v_{i-1}$ to $v_i$ in a topological map $\mathcal{T}$ following $\mathcal{W}$ toward $\tilde{v}_i$ generated by the action predictor $\pi$ with pose and depth observations $\{(\mathbf{p}, \mathbf{D})^*\}$. Since imminent collision is predicted by the collision predictor $\phi$ based on the current $\mathbf{D}$, $\tilde{v}_i$ is pose-corrected to $\hat{v}_i$ by $\pi$. Future SG estimates such as $\tilde{v}_{i+1}$ are also adjusted.}
    \label{fig:problem_formulation}
\end{figure}

\begin{figure*}[htb!]
\captionsetup{skip=0pt}
     \centering
     \includegraphics[width=0.95\linewidth]{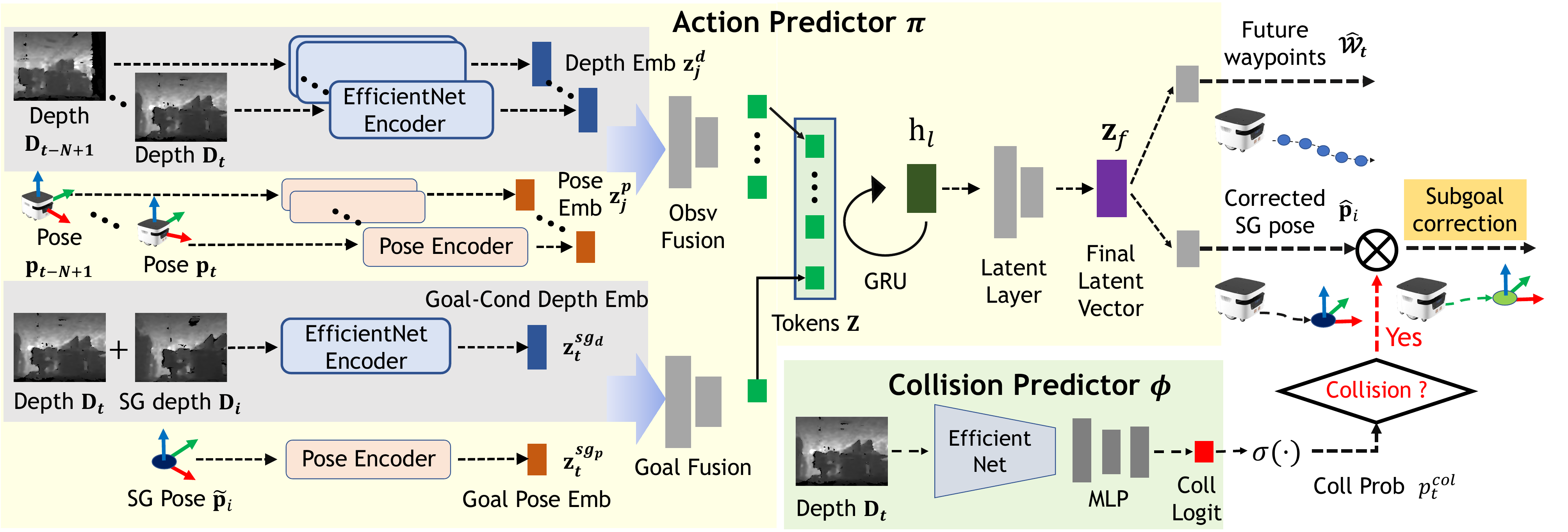}
     \caption{\label{fig:devgru_pipeline} 
     {{\bf DevGRU Architecture and Navigation Pipeline.} 
     }
     }
     \vspace*{-1.5em}
\end{figure*}

{
Our objective of visual navigation~\cite{ThrFox05}  for a mobile robot $\mathcal{R}$ over an environment $\Omega$ is to move from a start node $v_1\in \mathcal{V}$ to the goal node $v_M\in \mathcal{V}$ by traversing a sequence of   
nodes $v_i \in \mathcal{V}$ in a directed graph $\mathcal{T} = (\mathcal{V}, \mathcal{E})$, representing $\Omega$, using a depth image $\mathbf{D}_i$. 
Here, each node $v_i \in \mathcal{V}$ is defined as $v_i = (\mathbf{p}_i, \mathbf{D}_i)$, where $\mathbf{p}_i$ is a collision-free pose of $\mathcal{R}$
and $\mathbf{D}_i\in\mathbb{R}^{H \times W}$ is the depth image observed by $\mathcal{R}$ posed at $\mathbf{p}_i$, denoted as $\mathcal{R}(\mathbf{p}_i)$. Each directed edge $e_{ij} \in \mathcal{E}$ corresponds to a collision-free transition of $\mathcal{R}$ from $v_i$ to $v_j$ according to a policy $\pi: \mathcal{V}\times \mathcal{C}\rightarrow \mathcal{V}\times\mathcal{W}$ where $\mathcal{C}$ is {\em recent} observations on $\Omega$ and $\mathcal{W}$ is $\mathcal{R}$'s future trajectory within a finite horizon. $\mathcal{T}$ is known as a {\em topological map} of $\Omega$, and $v_{i\neq M} \in \mathcal{V}$ is also called a {\em sub-goal} (SG) node.
In practice, $\mathcal{T}$ is precomputed by manually navigating $\mathcal{R}$ over $\Omega$, where the nodes $\mathcal{V}$ and edges $\mathcal{E}$ are obtained by sampling the robot's trajectory at a certain interval\cite{ShaLev23vint}.
}
{
An SG node $v_i=(\mathbf{p}_i,*)$ is considered {\em reached} by $\mathcal{R}$ located at $\mathbf{x}$  if $\left\|{\mathbf{p}}_i - \mathbf{x}\right\|_{\mathcal{M}} < \epsilon_\mathcal{R}$ under some configuration-space metric $\mathcal{M}$ where  $\epsilon_\mathcal{R}$ is a distance threshold. 
}



\subsection{Framework Overview } 
\label{sec:overview}

{Our DevGRU framework is designed to perform topological navigation using a topological map $\mathcal{T}$ for $\mathcal{R}$, as illustrated in Fig.~\ref{fig:problem_formulation}. 
Since DevGRU does not use a metric map, $\mathcal{T}$ is estimated by  $\mathcal{R}$'s odometry during navigation, which can be inaccurate at times.
Starting from $v_1\in \mathcal{T}$, $\mathcal{R}$ moves from the SG $v_{i-1}\in \mathcal{T}$ to the next SG $v_i\in \mathcal{T}$ until reaching the goal node. 
DevGRU uses the Action Predictor (AP) $\pi$, which takes the target SG $v_i$ with past $N$ observations $\mathcal{C}$ of $\mathcal{R}$'s poses and observed depth images to generate a trajectory of 
waypoints ${\hat{\mathcal{W}}}$. 
By repeatedly executing $\pi$ and moving along ${\hat{\mathcal{W}}}$,  $\mathcal{R}$ attempts to reach ${v}_i$. However, $\tilde{v}_i$, $v_i$ seen from $\mathcal{R}$, can be deviated from $v_i$ due to the odometry error. As a result, $\tilde{v}_i$ can become inaccurate or even induce imminent collision.
Thus, $\pi$ also outputs $\hat{v}_i$, which is pose-corrected from $\tilde{v}_i$ to be close to $v_i$ and collision free.
When $\mathcal{R}$ reaches $\hat{v}_i$, the SG $v_i$ in $\mathcal{T}$ is considered reached. 
}
{Moreover, when a Collision Predictor (CP) projects a collision from the depth $\mathbf{D}_t$ observed by $\mathcal{R}$ at time $t$, DevGRU also corrects all the future sub-goal poses in $\tilde{\mathcal{V}}_{i}=\{\tilde{v}_j| \forall j \ge i\}$
by propagating the error offset between $\tilde{v}_i$ and $\hat{v}_i$  to the rest of SGs in $\tilde{\mathcal{V}}_{i}$, yielding $\hat{\mathcal{V}}_{i}$.
}
{
The following sections present a step-by-step description of the AP and CP architectures (also illustrated in Fig.~\ref{fig:devgru_pipeline}).
}
\subsection{Action Predictor} 
\label{sec:Action Predictor}
Assume that $\mathcal{R}$'s pose is $\mathbf{p}_t$ at a current discrete timestep $t\in \mathbb{R}^+$
and the target SG is $v_i$.
The objective of the AP $\pi$ is to generate a trajectory ${\hat{\mathcal{W}}}$ for $\mathcal{R}$ to reach $v_i$ from $v_{i-1}$ with {$N$ recent observations} $\mathcal{C}_t=\{( \mathbf{p}_{j},\mathbf{D}_j)|t-N+1\le \forall j \le t\}$ of $\mathcal{R}$'s poses $\mathbf{p}_j$ and depth images $\mathbf{D}_j$ observed by $\mathcal{R}(\mathbf{p}_j)$.

{\bf Architecture:}
Given a sequence of {$N$\footnote{{In our implementation, $N=6$.}}} depth images ${\mathcal{D}}=\{\mathbf{D}_{t-N+1:t}\}$, each frame in $\mathcal{D}$ is independently encoded by an EfficientNet Encoder composed of EfficientNet-B0 backbone \cite{TanLe19} followed by a linear projection, producing a depth embeddings $\mathbf{z}^d_j$. In parallel, the corresponding per-frame pose inputs $\mathbf{p}_j$ for $\mathcal{R}$ are encoded by an MLP-based Pose Encoder\footnote{{Linear$(\{2\vert4\} \rightarrow 128) $, ReLU, Linear$(128 \rightarrow 64)$} }, yielding pose embeddings $\mathbf{z}^p_j$.
The depth and pose embeddings are concatenated $[\mathbf{z}^d_j ; \mathbf{z}^p_j]$ and passed through a fusion MLP\footnote{{Linear$(576 \rightarrow 512)$, ReLU, Dropout} }, producing fused observation tokens $\mathbf{z}^o_j$.

In addition to the observation tokens, we construct a goal token corresponding to the target SG $v_i$. Concretely, the current depth image $\mathbf{D}_t$ and the SG depth image $\mathbf{D}_i$ from $v_i$ are concatenated and processed by the EfficientNet Encoder, producing a goal-conditioned depth embedding $\mathbf{z}^{sg_d}_t$. Simultaneously, the SG pose $\mathbf{p}_{i}$ in $v_i$ is encoded by the Pose Encoder into a goal pose embedding $\mathbf{z}^{sg_p}_t$. The two embeddings are then concatenated and passed through a fusion MLP to obtain the fused goal token $\mathbf{z}^{sg}_t$. Then, the fused observation and goal token are concatenated to form the input sequence
\[
\mathbf{Z} =
\{\mathbf{z}^o_{t-N+1}, \dots, \mathbf{z}^o_t, \mathbf{z}^{sg}_t\}
\]
This sequence is processed by a GRU\cite{ChoBen14emnlp}, which iteratively updates a hidden state to capture temporal dependencies across the observation history and the goal token. Specifically, at each step $l$,
\[
\mathbf{h}_l = \text{GRU}(\mathbf{z}_l, \mathbf{h}_{l-1}),
\]
where $\mathbf{z}_l \in \mathbf{Z}$ denotes the $l$-th input token, and $\mathbf{h}_{l-1}$ and $\mathbf{h}_l$ denote the hidden states. The last two hidden states are concatenated and projected through an MLP to obtain the final latent vector $\mathbf{z}_f$.
 $\mathbf{z}_f$ is used by subsequent prediction heads to estimate the collision-aware trajectory of waypoints $\hat{\mathcal{W}}$ as well as {correct} the SG pose $\mathbf{\hat{p}}_i$ in $\hat{v}_i$.

{\bf Training Loss:}
The AP network is trained with a multi-task objective that jointly supervises the SG pose and a sequence of future waypoints $\hat{\mathcal{W}}=\{\hat{\mathcal{W}}_k|t+1 \le k \le t+K\}$ where $K$\footnote{In our implementation, $K=5$.} is the prediction horizon. 
All position components in the SG and waypoints are normalized by a maximum distance per waypoint before training to balance the position and orientation terms in the loss. 

{Overall, the total loss of AP is defined as the joint minimization of waypoint loss $\mathcal{L}_{wp}$ to predict $\hat{\mathcal{W}}$ and SG pose loss $\mathcal{L}_{sg}$ to predict $\mathbf{\hat{p}}$ in $\hat{v}$:}
\begin{equation}
\mathcal{L}_\psi =
\alpha \mathcal{L}_{wp} +
(1-\alpha)\mathcal{L}_{sg},
\end{equation}
where $\alpha \in [0,1]$ balances trajectory and SG pose supervision\footnote{$\alpha$ is set to $0.5$ in our implementation.}.
{$\mathcal{L}_{sg}$} consists of a translational loss---defined as the sum of Smooth L1 losses $\mathcal{H}(\cdot)$ in the 2D plane---
and an orientation loss:
\begin{equation}
\mathcal{L}_{sg} =
\mathcal{H}(\hat{\mathbf{x}}_{sg},\mathbf{x}_{sg})
+ \left(1 - \langle \hat{\mathbf{q}}_{sg}, \mathbf{q}_{sg} \rangle^2\right),
\end{equation}
where $\hat{\mathbf{x}}_{sg}, \mathbf{x}_{sg} \in \mathbb{R}^2$ denote the predicted
and ground-truth SG positions in normalized coordinates, respectively. $\hat{\mathbf{q}}_{sg}, \mathbf{q}_{sg} \in \mathbb{H}$ denote predicted and ground-truth SG orientations in unit quaternions. $\langle \cdot,\cdot \rangle$ denotes the standard inner product on $\mathbb{H}$.
Similar to SG pose prediction, each normalized waypoint pose is decomposed into position $\mathbf{x}_k\in \mathbb{R}^2$ and orientation $\mathbf{q}_k$. Accordingly, {$\mathcal{L}_{wp}$} minimizes the error with respect to the ground-truth waypoints:
\begin{equation}
\mathcal{L}_{wp} =
\frac{1}{K}\sum_{k=1}^{K}
\left\|\hat{\mathbf{x}}_k - \mathbf{x}_k\right\|_2^2
+
\frac{1}{K}\sum_{k=1}^{K}
\left(1 -  \langle \hat{\mathbf{q}}_k, \mathbf{q}_k \rangle ^2 \right).
\end{equation}


\subsection{Collision Predictor} 
\label{collision_predictor}
Aside from the action predictor model, we have trained a depth-based binary classifier that maps a single normalized depth observation to a collision likelihood. A depth image at frame $t$ is encoded using an EfficientNet backbone to a feature representation, which is projected to a linear layer, producing $\mathbf{z}^d_t$ 
embedding, followed by SiLU activation and dropout. A linear classification head then produces a scalar collision logit $s\in \mathbb{R}$, which is converted to a collision probability by a sigmoid function $p^{col}=(1+\exp(-s))^{-1}$.

\subsection{Collision-aware Navigation} 
\label{devgru_deployment}
{The outputs of the AP $\pi$ and CP $\phi$ provide navigation information at time $t$, including collision-aware waypoints $\hat{\mathcal W}_t$, corrected SG poses $\hat{\mathbf{p}}$, and collision probabilities $p^{col}_t$. These outputs are jointly integrated to generate optimal actions, such as producing control commands, and adjusting future SG poses to proactively avoid potential collisions. The corresponding algorithm in DevGRU is described in the following:
}

\begin{enumerate}

\item[0)] {\bf Precomputation:} Build a topological map $\mathcal{T}$ from the environment $\Omega$.
\item {\bf Initialization:} Set the current SG $v_{i\leftarrow1}$ and $\mathcal{R}$'s pose $\mathbf{p}_{t\leftarrow t_1}$ at the initial time $t_1$.

\item Repeat a)$\sim$b) until $\mathcal{R}(\mathbf{p}_t)$ {\em reaches} $v_M$:
\begin{enumerate}

\item Repeat i)$\sim$iv) until $\mathcal{R}(\mathbf{p}_t)$ {\em reaches} $v_i=(\mathbf{p}_i,\mathbf{D}_i)$: 


\begin{enumerate}

\item {\bf Robot Pose Estimation:} Estimate $\mathbf{p}_i$ to $\tilde{\mathbf{p}}_i$ using  $\mathcal{R}$'s odometry, and set $\tilde{v}_i = (\tilde{\mathbf{p}}_i, \mathbf{D}_i)$.

\item {\bf Action Prediction:} Aggregate the $N$ observation context
$\mathcal{C}_t=\{\mathbf{O}_{t-N+1:t}\}, \mathbf{O}_t=({\mathbf{p}_t}, \mathbf{D}_t)$ with the {depth image} $\mathbf{D}_t$ observed by $\mathcal{R}(\mathbf{p}_t)$
and perform AP inference: 
\[
(\hat{{v}}_i, \hat{\mathcal{W}}_t)
\gets
\pi(\tilde{v}_i, \mathcal{C}_t)
\]
{where ${\hat{v}}_i$ and $\hat{\mathcal{W}}_t$ denote the pose-corrected SG and the waypoints, respectively.}


\item {\bf SG Pose Correction:} Predict the collision likelihood from the {observed depth image}, 
${p^{\text{col}}_{t}} = \phi(\mathbf{D}_t)$, and if $p^\text{col}_{t}$ is sufficiently high, correct the future SGs' poses:\label{pseudo-code:sg_correction}
\[
\hat{\mathbf{p}}_{i:M}
\leftarrow
\tilde{\mathbf{p}}_{i:M} \oplus \Delta{\mathbf{p}},
\qquad
\Delta{\mathbf{p}}=
\hat{\mathbf{p}}_i \ominus \tilde{\mathbf{p}}_i.
\]
where $\hat{\mathbf{p}}_i$ is the pose component of $\hat{v}_i$ and $\oplus$ and $\ominus$ are the pose composition operators in the configuration space of $\mathcal{R}$~\cite{GriBur10}.

\item {\bf Action:} Feed $\hat{\mathcal{W}}_t$ to a position controller for $\mathcal{R}$ to move to $\mathbf{p}_{t+\Delta t\leftarrow t}$.


\end{enumerate}

\item Set $i$ to $j$ when ${e}_{ij} \in \mathcal{E}$ of $\mathcal{T}$.


\end{enumerate}
\end{enumerate}

\section{Training Data Collection}
\label{sec:data_collection}

{Unlike many recent embodied AI navigation datasets~\cite{HirSav19}\cite{KahLev18}\cite{TanLiu24}, our dataset explicitly includes collision events encountered during real-robot navigation. This balanced training data consists of both collision and collision-free events. Accordingly, the robot is exposed to diverse collision scenarios, enabling it to acquire collision-aware navigation skills while maintaining fundamental free-space navigation capability. 
} 

\subsection{Collision-free Navigation Data}
\label{subsec:non-collision_data}
\begin{figure}[htb!]
\centering
\setlength{\tabcolsep}{0.5pt}
\setlength{\fboxsep}{0.3pt}

\begin{tabular}{cccc}
\includegraphics[width=0.235\columnwidth]{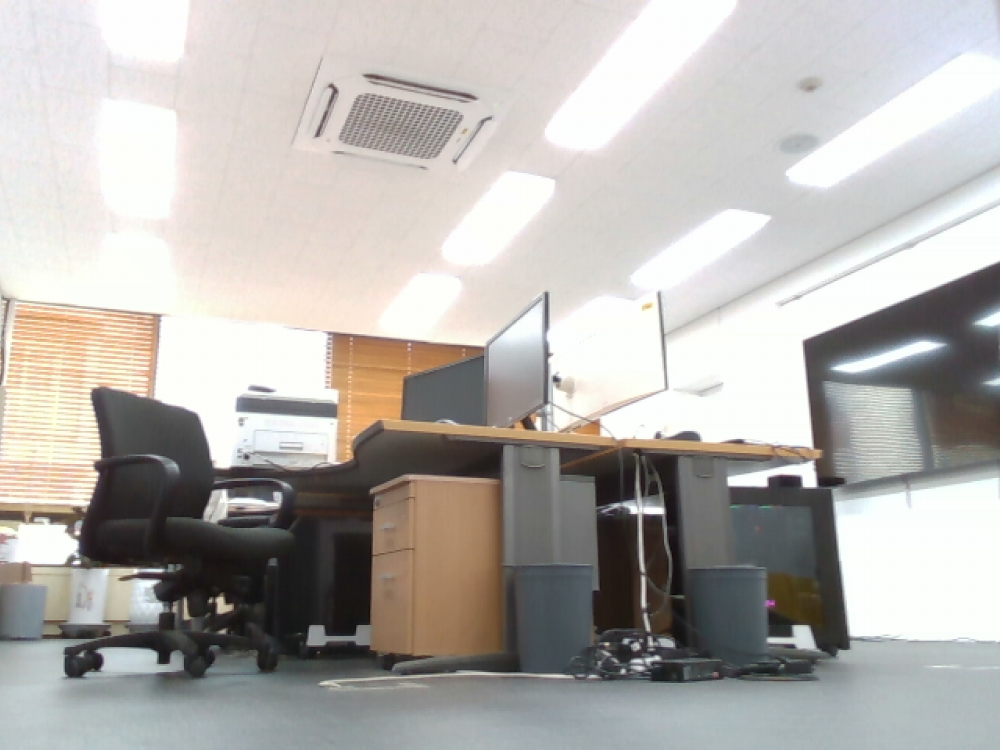} &
\includegraphics[width=0.235\columnwidth]{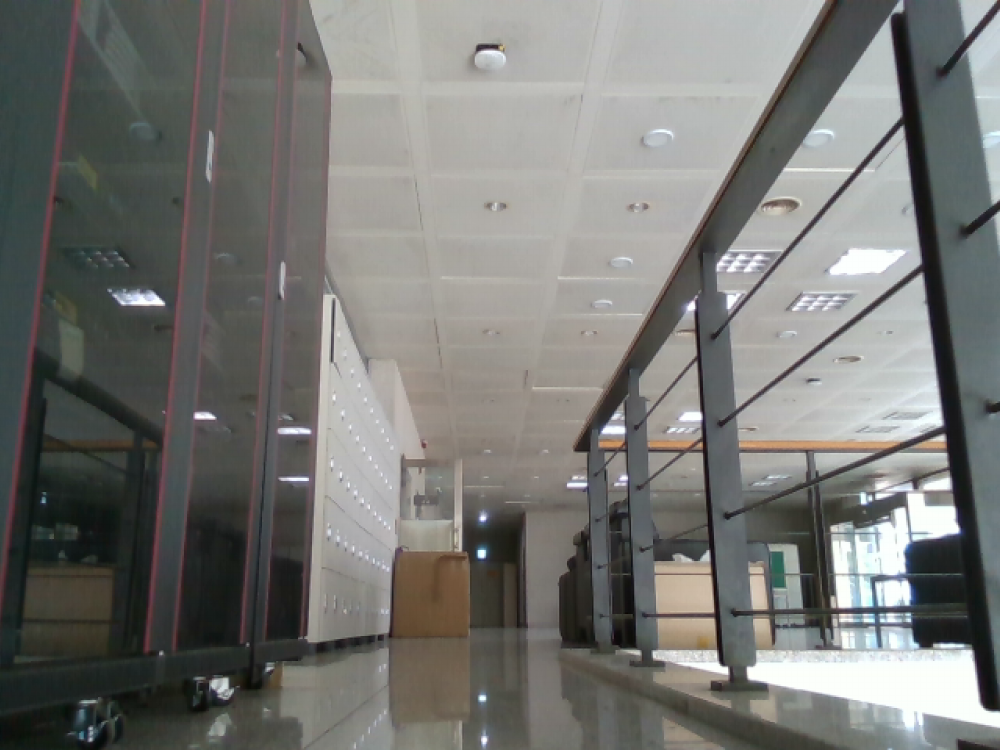} &
\includegraphics[width=0.235\columnwidth]{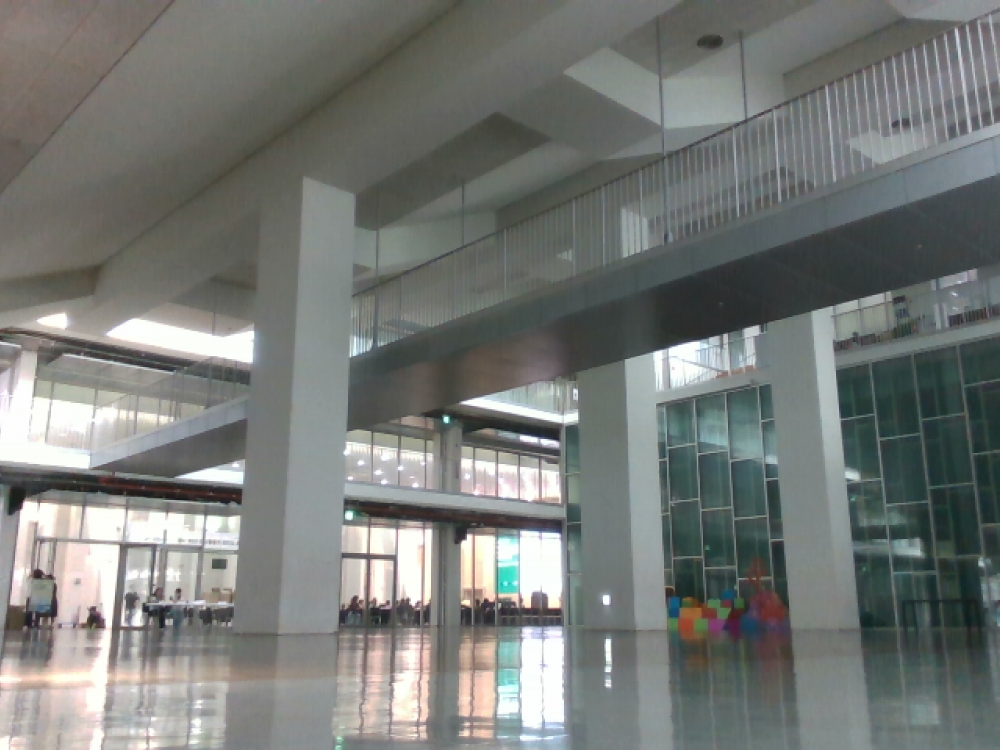} &
\includegraphics[width=0.235\columnwidth]{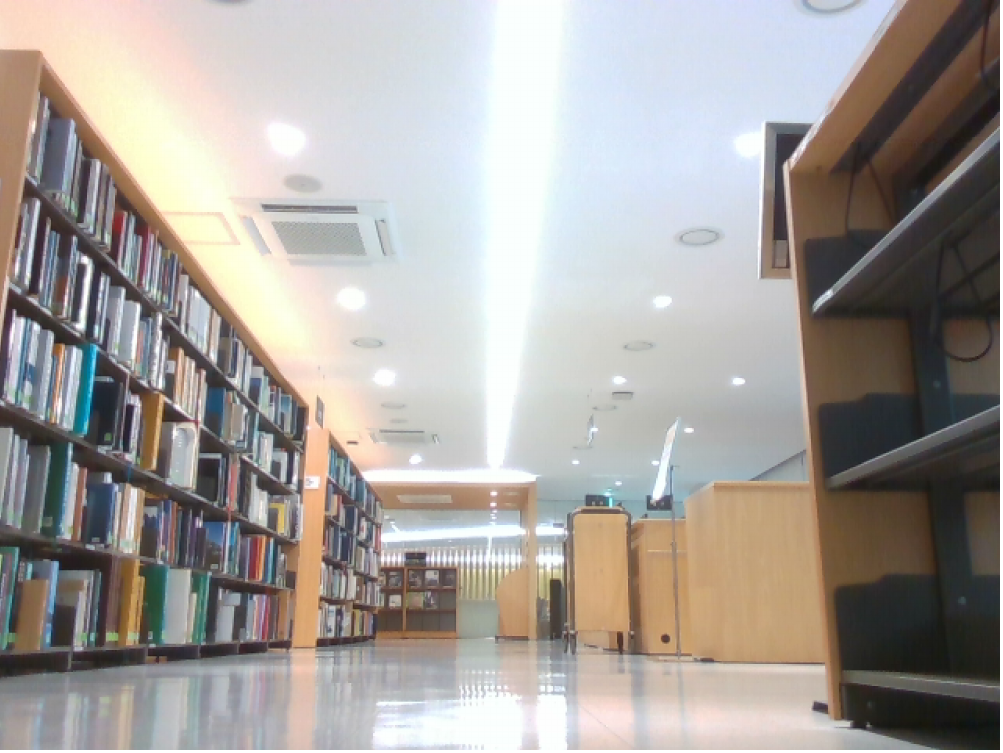} \\[-2pt]

\includegraphics[width=0.235\columnwidth]{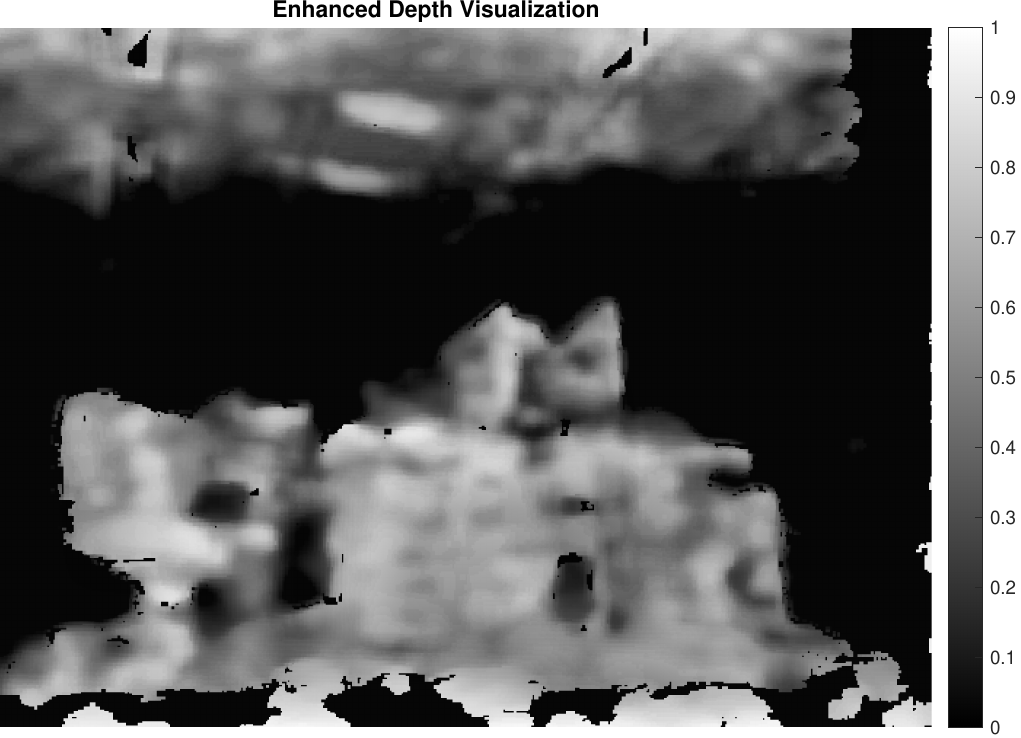} &
\includegraphics[width=0.235\columnwidth]{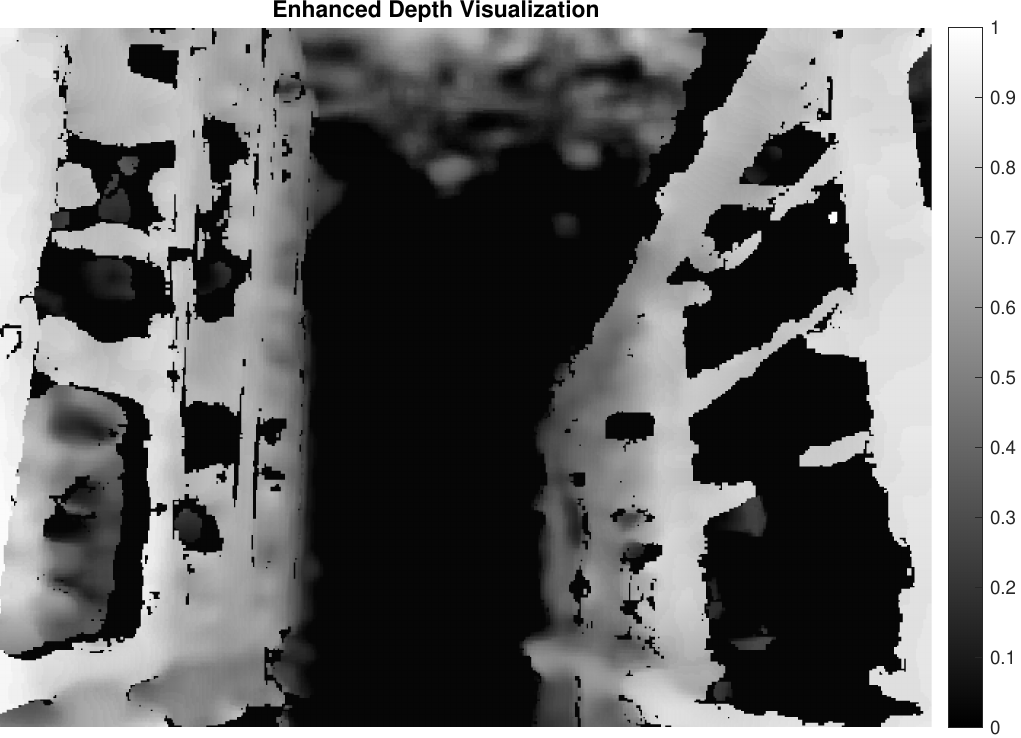} &
\includegraphics[width=0.235\columnwidth]{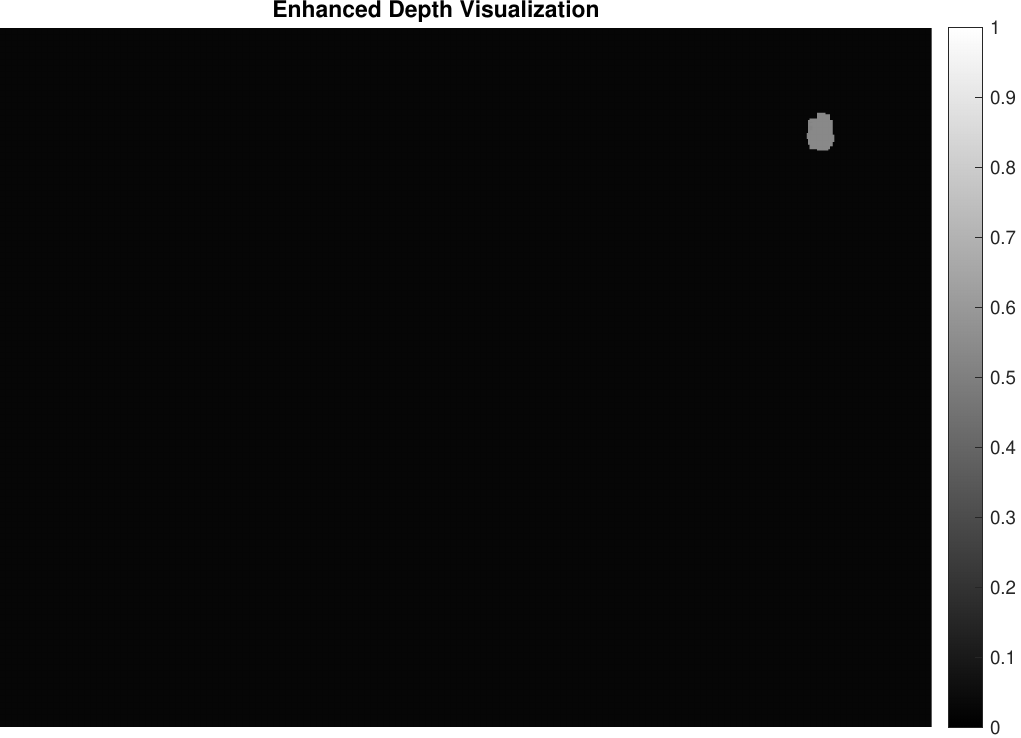} &
\includegraphics[width=0.235\columnwidth]{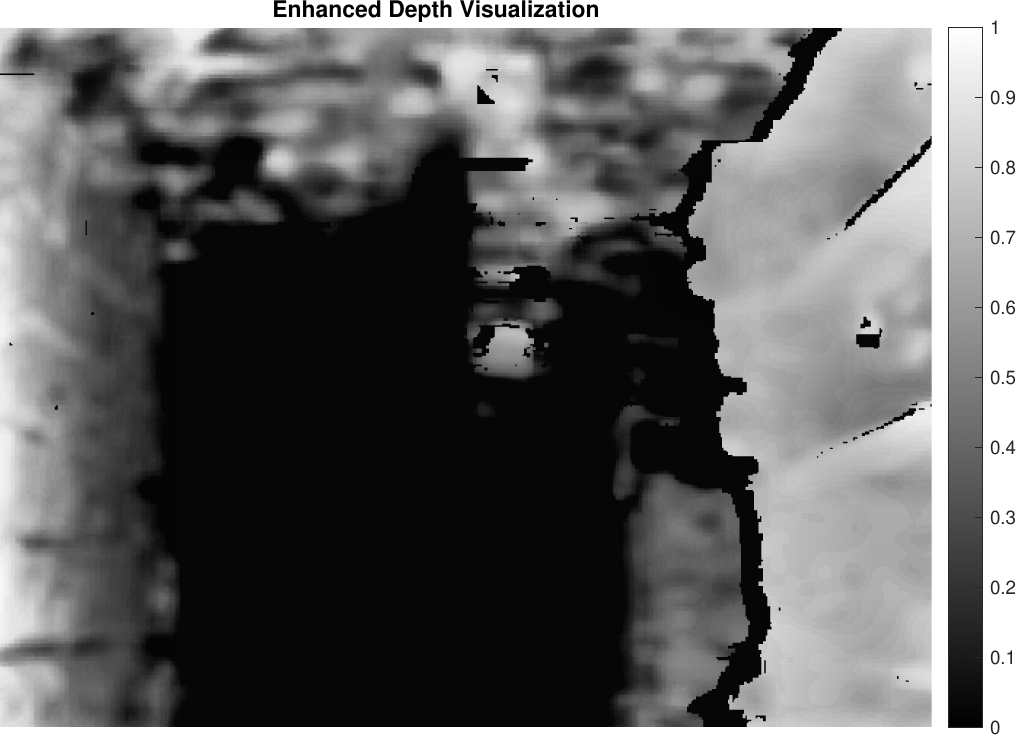} \\[-1pt]

\scriptsize Office &
\scriptsize Building A &
\scriptsize Building B &
\scriptsize Library
\end{tabular}

\caption{Example RGB images (top row) and corresponding depth images (bottom row) collected from various training sites that are distinct from the evaluation environments.}
\label{fig:sample_rgb_depth_pairs}
\end{figure}

To construct the {collision-free} dataset, we visited five buildings across the university campus. In each building, a human operator manually teleoperated the robot to explore diverse indoor areas, including corridors, classrooms, and hallways; see Fig~\ref{fig:sample_rgb_depth_pairs} for sample images. 
During teleoperation, we recorded  {time-synchronized depth images and the robot poses} estimated from wheel odometry. 
From the teleoperated runs, we collected approximately 10K {collision-free} navigation samples for training. Each sample at time $t$ consists of a sequence of six depth observations $\mathbf{D}_{t-5:t}$ (the current and five previous frames), the corresponding robot odometry poses $\mathbf{p}_{t-5:t}$, {a subgoal pose $\mathbf{p}$,} 
and a sequence of five future waypoints $\mathcal{W}_t$ with respect to the current robot pose $\mathbf{p}_t$. Each sample is represented as a tuple:
$[\mathbf{D}_{t-5:t},\ \mathbf{p}_{t-5:t},\, \mathbf{p},\, \mathcal{W}_t]$

\subsection{Collision Navigation Data}\label{subsec:collision_data_collection}

{The AP policy $\pi_0$, trained exclusively on a collision-free dataset, may be vulnerable to collision events. To address this limitation, we construct a collision-intensive dataset by reproducing collision events caused by the original policy $\pi_0$. These reproduced failures are used to augment the dataset with both collision trajectories. A new AP, $\pi_1$, is then trained on the combined dataset, resulting in greater robustness than $\pi_0$. To obtain this goal, we use a semi-automatic data collection strategy.}

\begin{enumerate}
\item We trained the AP $\pi_0$ on the collision-free dataset obtained from Section~\ref{subsec:non-collision_data}, yielding the \textit{DevGRU-Base} model. 
\item We deployed {DevGRU-Base} in 
the same environments used for collecting collision-free dataset using $\pi_0$ but without using the CP $\phi$ step-\ref{pseudo-code:sg_correction} in Section~\ref{devgru_deployment}, and a human operator recorded all collision events.
\item For each recorded collision event, we recompute a new {collision-free} waypoint sequence $\mathcal{W}_t'$ and {the collision-free (corrected)} SG pose $\mathbf{p}'$ using $\mathcal{T}$. 
The recomputed $\mathcal{W}_t'$ and $\mathbf{p}'$ are used as new ground-truth targets for training the AP model $\pi_1$.
\end{enumerate}
Following this procedure, we collected approximately 10K collision navigation samples; 
{each sample follows the same tuple format described in Section~\ref{subsec:non-collision_data}, except where $\mathbf{p}'$ and $\mathcal{W}'_t$ are recomputed according to the procedure above.}

\begin{figure}[htb]
\centering
\includegraphics[width=0.75\columnwidth]{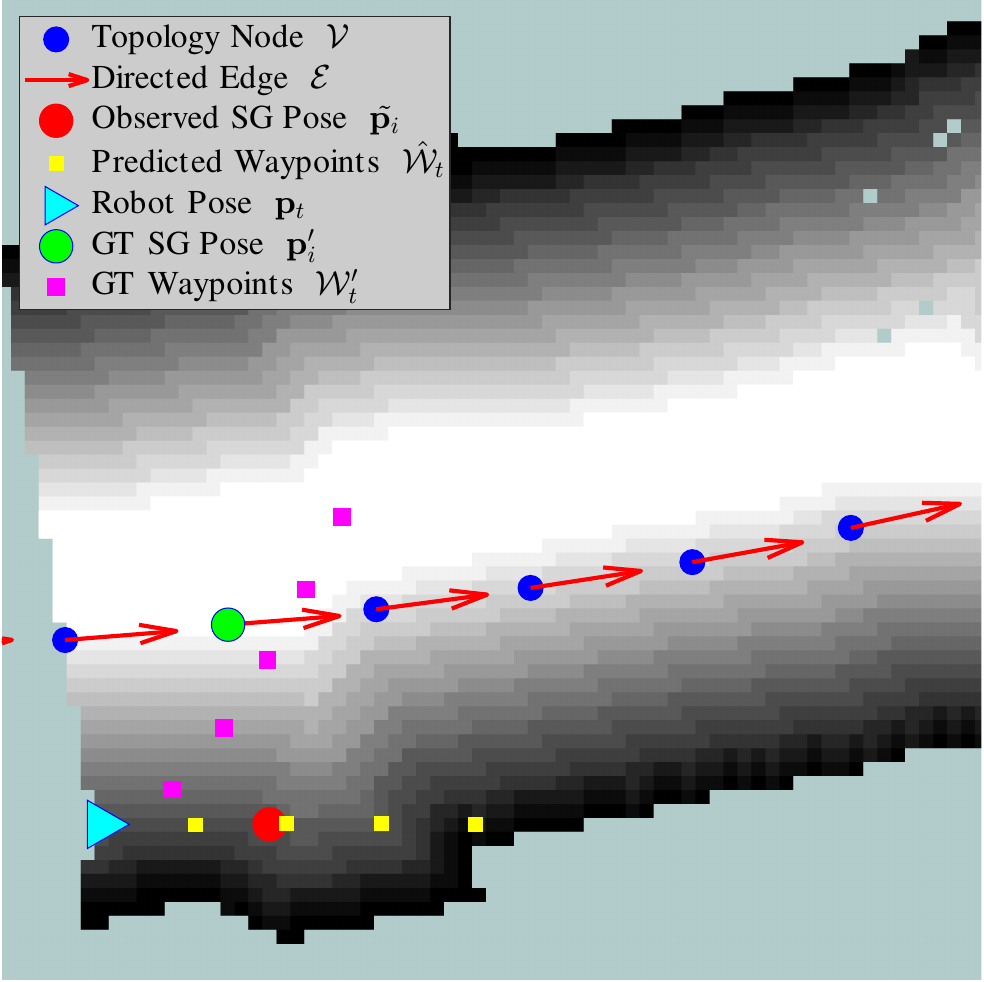}

\caption{
\textbf{Collision dataset configuration.}
The topological map is superimposed on top of the occupancy grid map. 
}
\label{fig:collision_config}
\end{figure}

Figure~\ref{fig:collision_config} illustrates an example configuration of a collision event. In this example, the robot approaches the right wall because it is guided by an error-accumulated SG pose $\tilde{\mathbf{p}}$ ({red solid circle}), while the corrected SG pose $\mathbf{p}$ is indicated by the green solid circle. 
Based on the corrected SG pose, we generate a new collision-free waypoint sequence $\mathcal{\mathcal{W}}_t$ ({magenta squares}).


{\bf Constrained Collision-Free Waypoint Generation}
\label{subsec:constrained_collision_free_data_generation}
The new waypoint sequence $\mathcal{W}_t'$ should be constrained to a collision-free trajectory, avoiding sharp turns and infeasible motions.
To satisfy this constraint, we first generate robot trajectories with bounded curvature by interpolating successive robot poses using cubic Hermite splines.
{Subsequently, the waypoint $\mathcal{W}_t'$ is adjusted based on the trajectory curvature, resulting in denser sampling in high-curvature regions and sparser sampling in low-curvature regions.}  
\section{Experiments and Results}
\label{Experiments and Results}

\subsection{Experiment Setup}
\label{subsec:experiment_setup}
\begin{figure}[htb!]
\centering
\includegraphics[width=0.5\columnwidth]{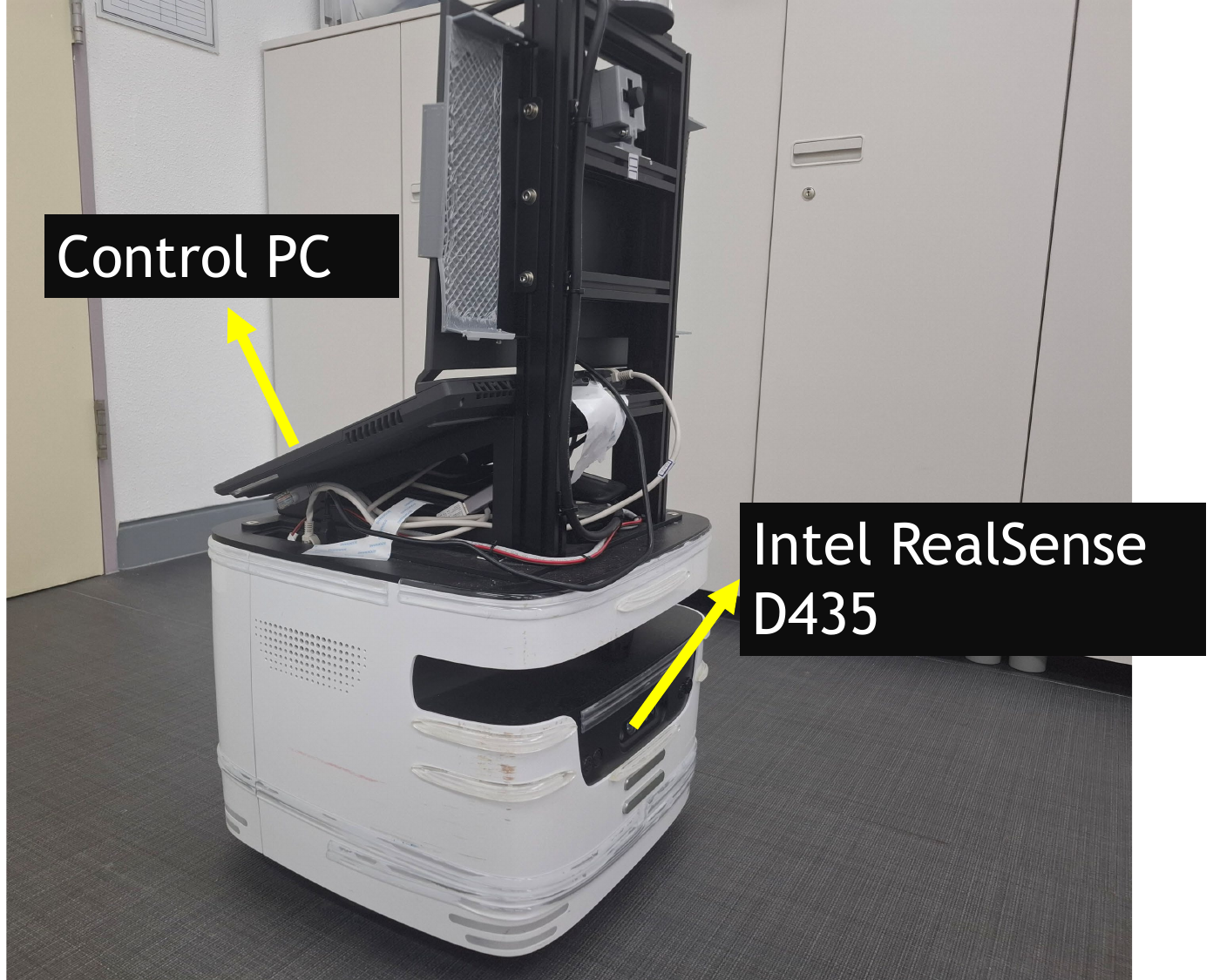}
\caption{The robot and its driver PC used for the experiments. 
}
\label{fig:former_robot}
\end{figure}

In the experiments, we employed a wheeled mobile robot equipped with an Intel RealSense D435 depth camera and a SICK TiM571 2D lidar sensor (see Figure \ref{fig:former_robot}). 
The 2D lidar sensor is used to collect navigation data and to perform post-analysis of each navigation run, but not for online navigation. On top of the robot, we mounted a driving laptop PC equipped with an AMD Ryzen 9 6900HX (1.6GHz) CPU and an Nvidia RTX 3070 GPU. All baseline models are integrated into ROS. 

{\bf Evaluation metrics:}
To quantify navigation performance, we employed the conventional Success weighted by Path Length (SPL) metric~\cite{AndZam18}. An episode is considered successful if the robot reaches the goal node $v_M\in \mathcal{T}$ from the start node $v_1\in \mathcal{T}$ within a predefined time limit, without collision. 
However, even if the agent has already explored a substantial portion of the topological map, an episode terminated by a collision renders SPL meaningless. Thus, we introduce node coverage (NC) {metric}, complementing SPL by quantifying the extent of {navigation} achieved during the episode, defined as
\begin{equation}
\mathrm{NC}
=
\frac{|\mathcal{V}_{\mathrm{vis}}|}{|\mathcal{V}|},
\end{equation}
where
$\mathcal{V}_{\mathrm{vis}}
=
\left\{
\forall v \in \mathcal{V}
\mid
\min_{\tau \in [0,T]}
\|\mathcal{R}(\tau) - \mathbf{p}_v\|
\le \epsilon_\mathcal{R}
\right\}
$
represents the set of nodes whose nodal position {$\mathbf{p}_v$} to the robot trajectory $\mathcal{R}(\tau)$ is set apart within $\epsilon_\mathcal{R}$ at any time during the navigation episode $[0,T]$. 

\begin{figure*}[!t]
\centering

\begin{subfigure}[c]{0.48\textwidth}
\centering
\includegraphics[width=\linewidth]{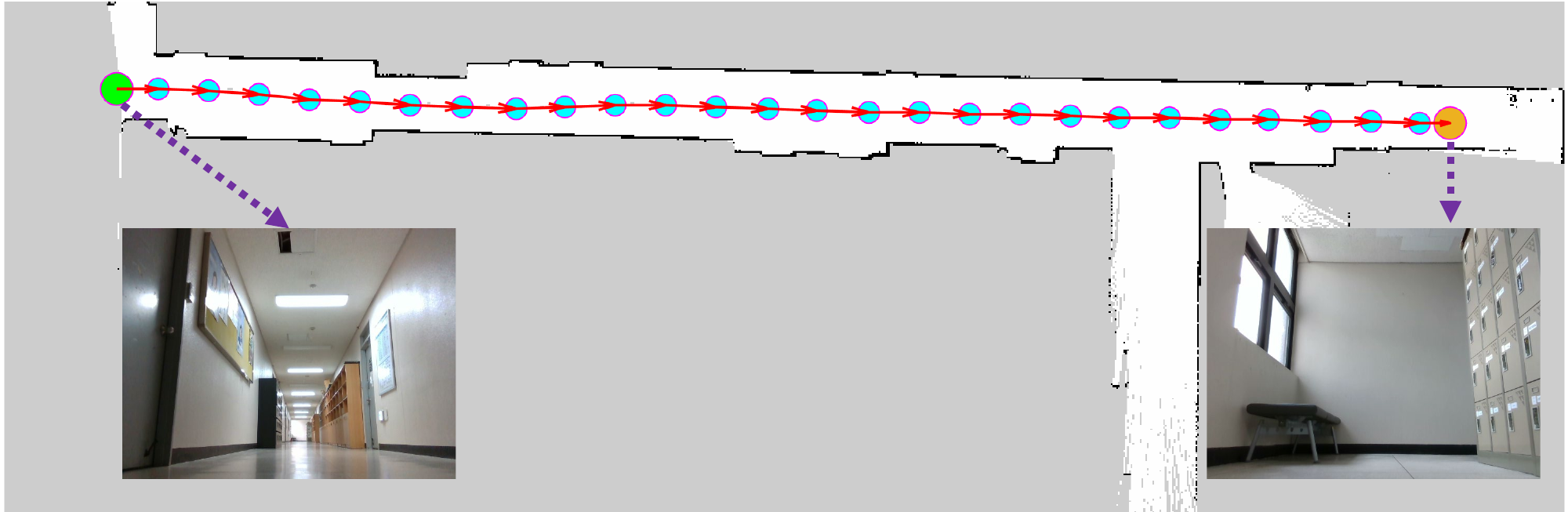}
\caption{T-B1}
\end{subfigure}
\hfill
\begin{subfigure}[c]{0.48\textwidth}
\centering
\includegraphics[width=\linewidth]{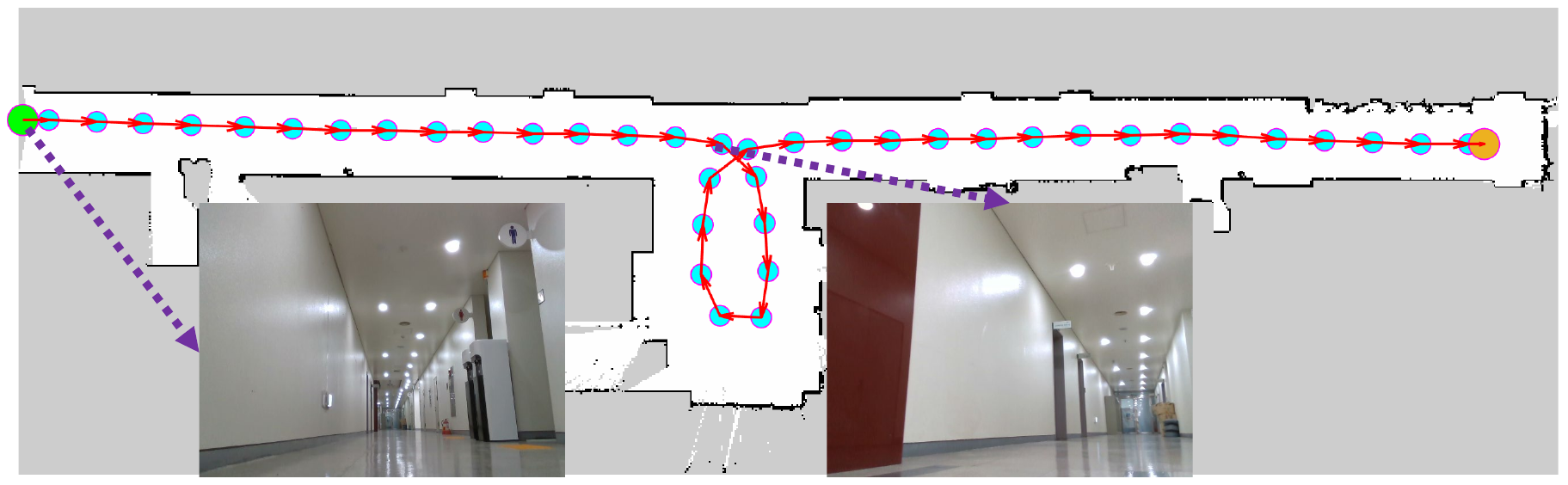}
\caption{T-B2}
\end{subfigure}
\\[-0.5mm]

\begin{subfigure}[c]{0.48\textwidth}
\centering
\includegraphics[width=\linewidth]{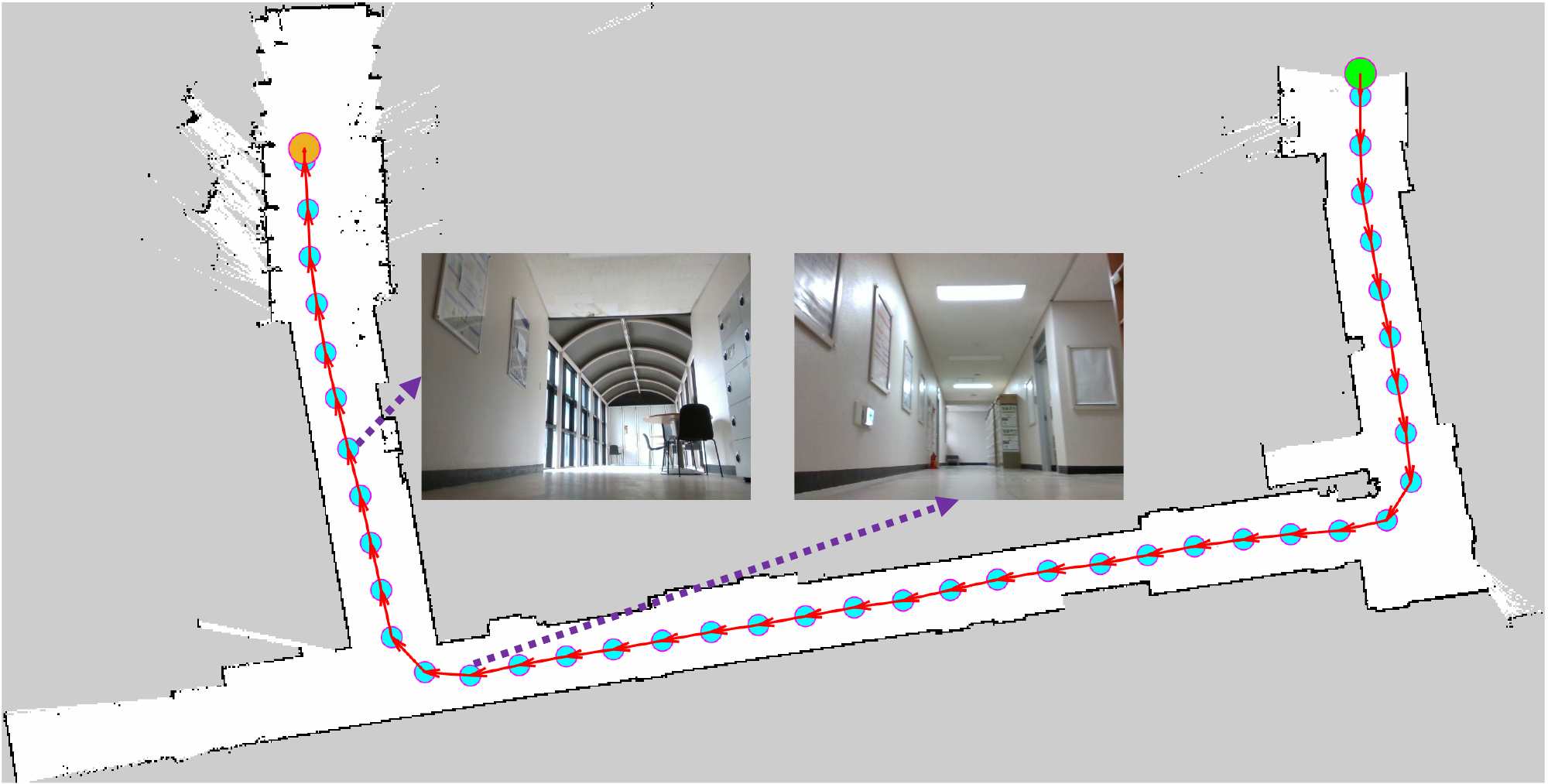}
\caption{T-B3}
\end{subfigure}
\hfill
\begin{subfigure}[c]{0.48\textwidth}
\centering
\includegraphics[width=\linewidth]{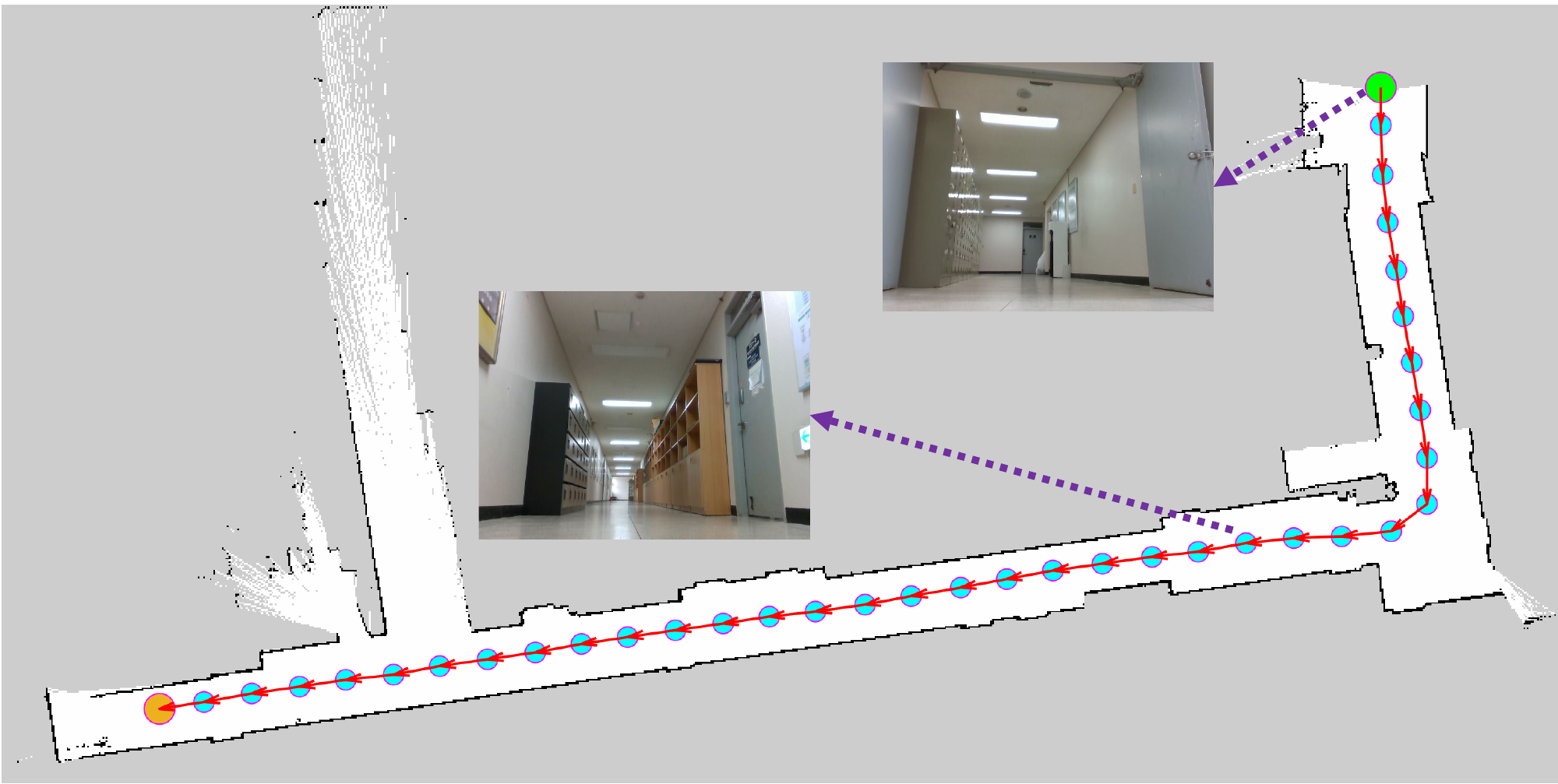}
\caption{T-B4}
\end{subfigure}
\\[1.5mm]

\begin{subfigure}[c]{0.48\textwidth}
\centering
\includegraphics[width=\linewidth]{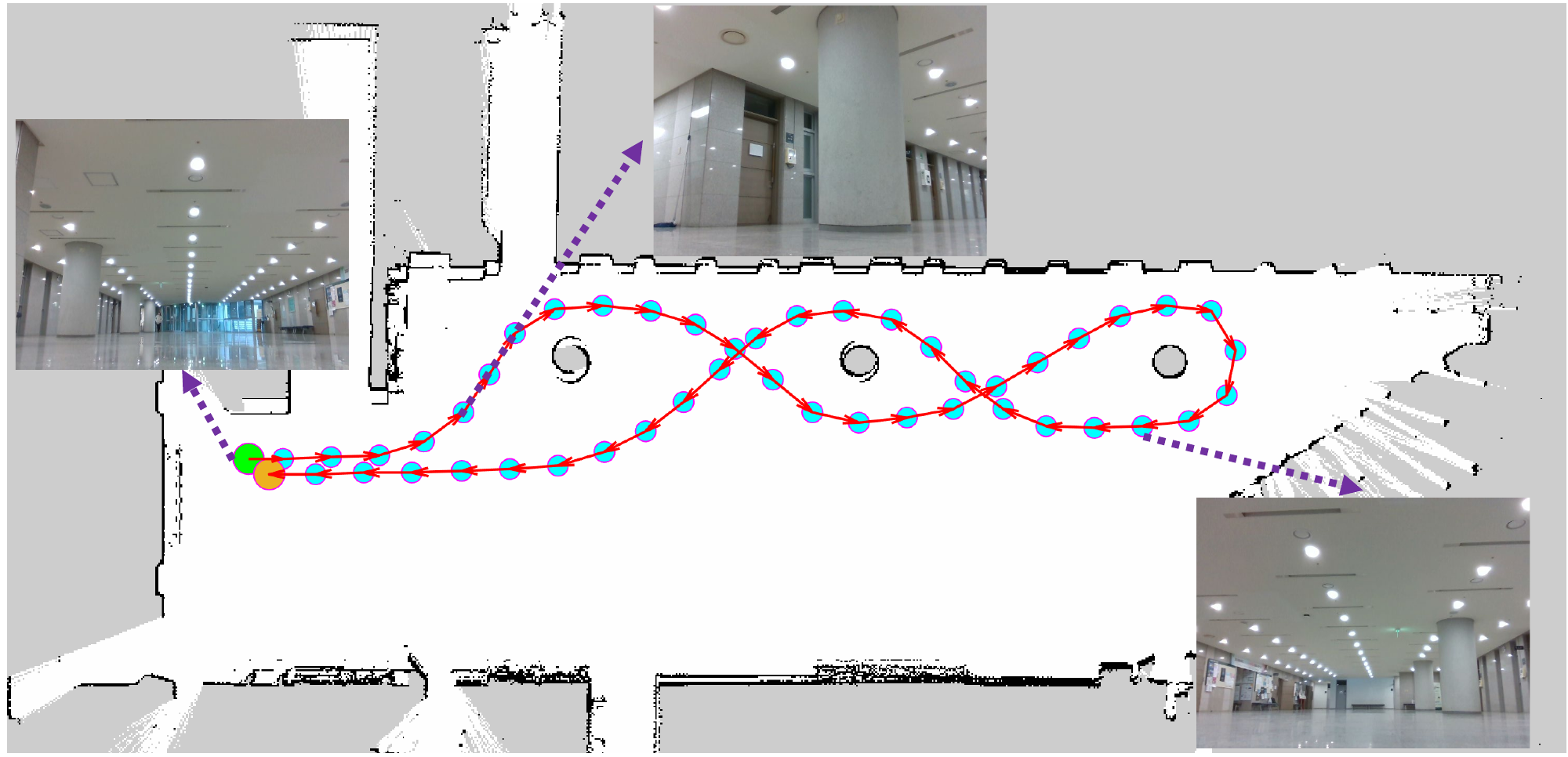}
\caption{T-B5}
\end{subfigure}
\hfill
\begin{subfigure}[c]{0.48\textwidth}
\centering
\includegraphics[width=\linewidth]{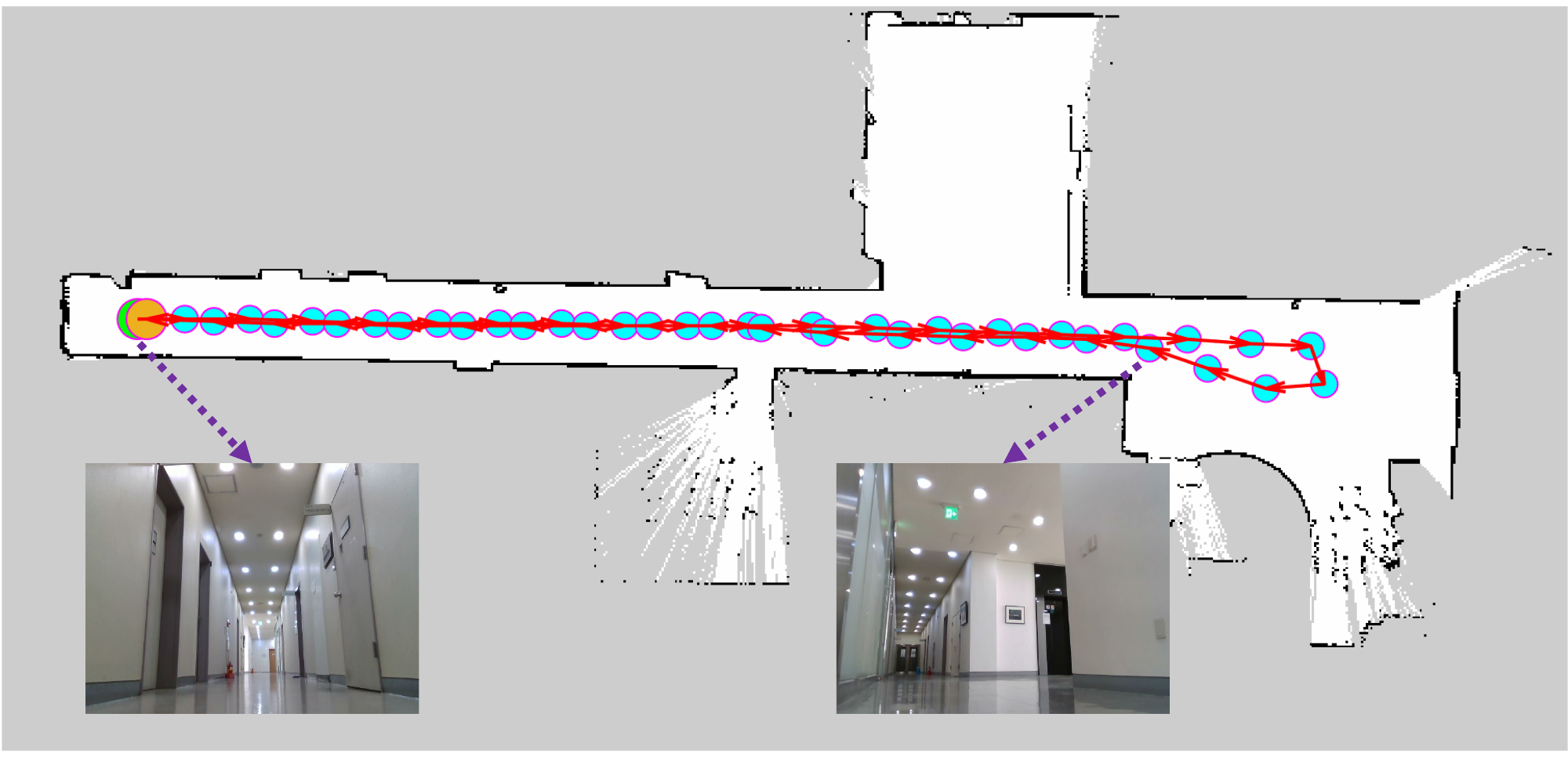}
\caption{T-B6}\label{fig:T-B6}
\end{subfigure}
\\[1.5mm]

\begin{subfigure}[c]{0.34\textwidth}
\centering
\setlength{\fboxsep}{0.3pt}
\setlength{\fboxrule}{0.5pt}

\includegraphics[width=\linewidth]{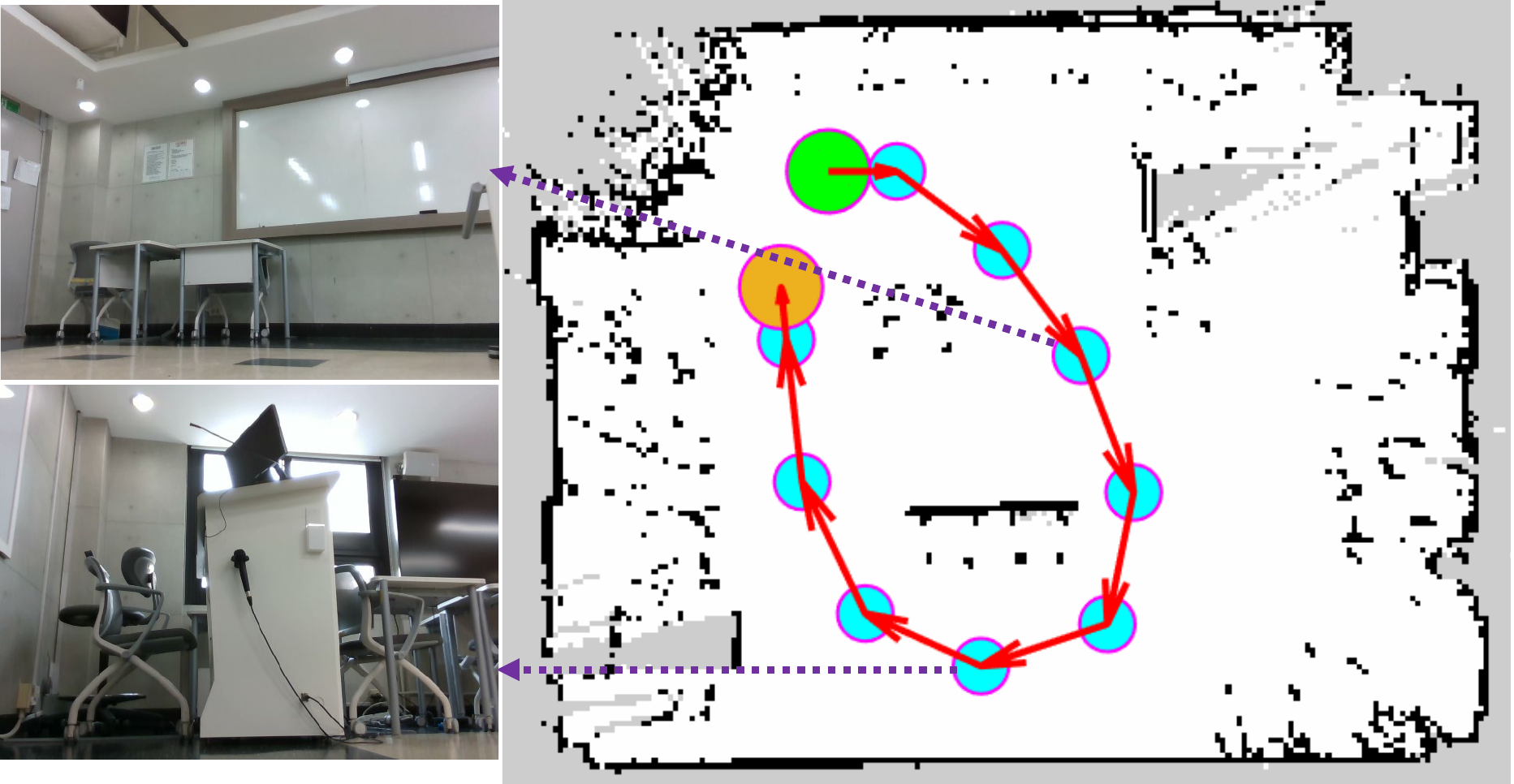}
\caption{T-B7}
\end{subfigure}
\hspace{0.2mm} 
\begin{subfigure}[c]{0.34\textwidth}
\centering
\setlength{\fboxsep}{0.3pt}
\setlength{\fboxrule}{0.5pt}

\includegraphics[width=\linewidth]{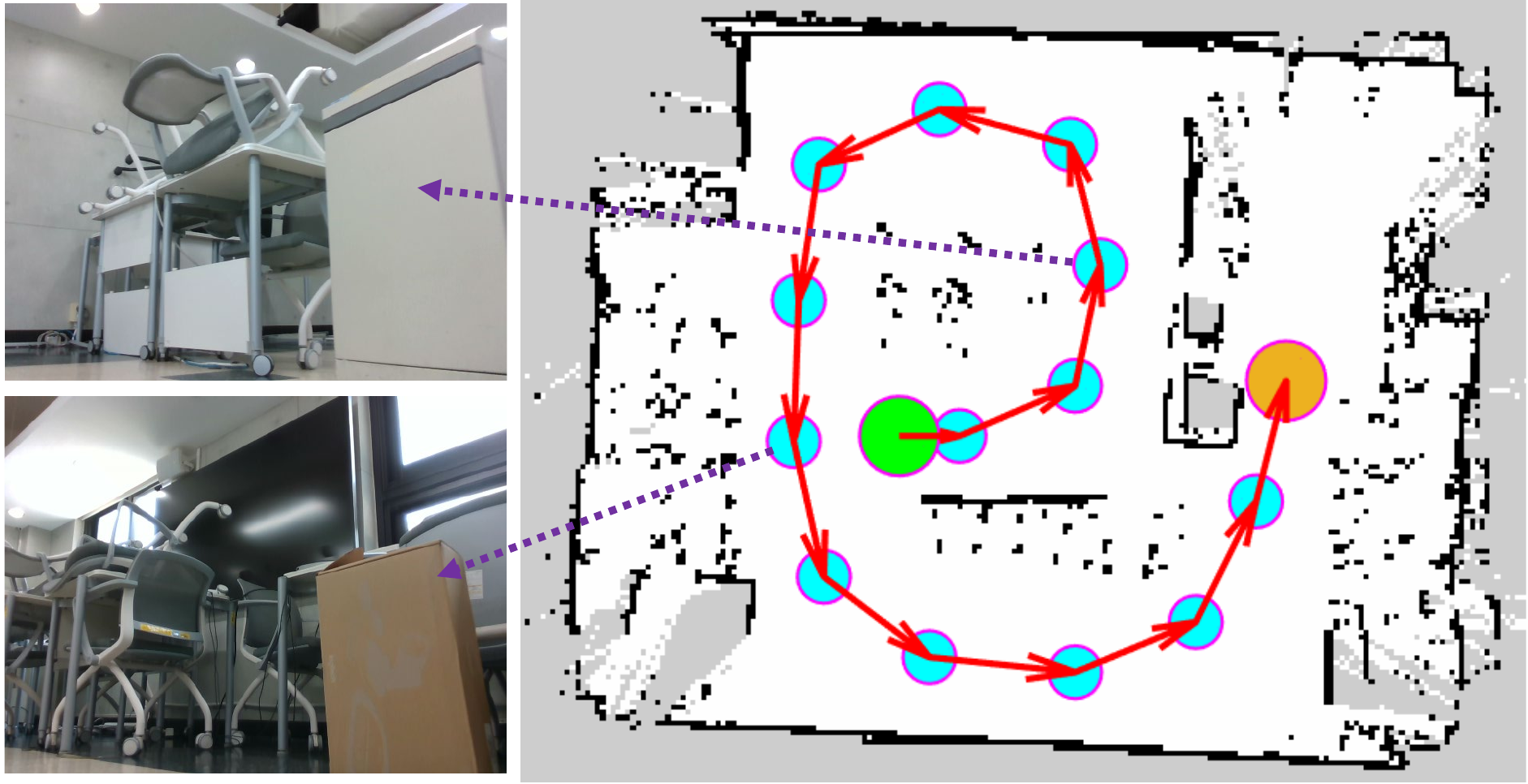}
\caption{T-B8}\label{fig:T-B8}
\end{subfigure}
\hspace{0.2mm} 
\begin{subfigure}[c]{0.27\textwidth}
\centering
\setlength{\fboxsep}{0.3pt}
\setlength{\fboxrule}{0.5pt}

\includegraphics[width=\linewidth]{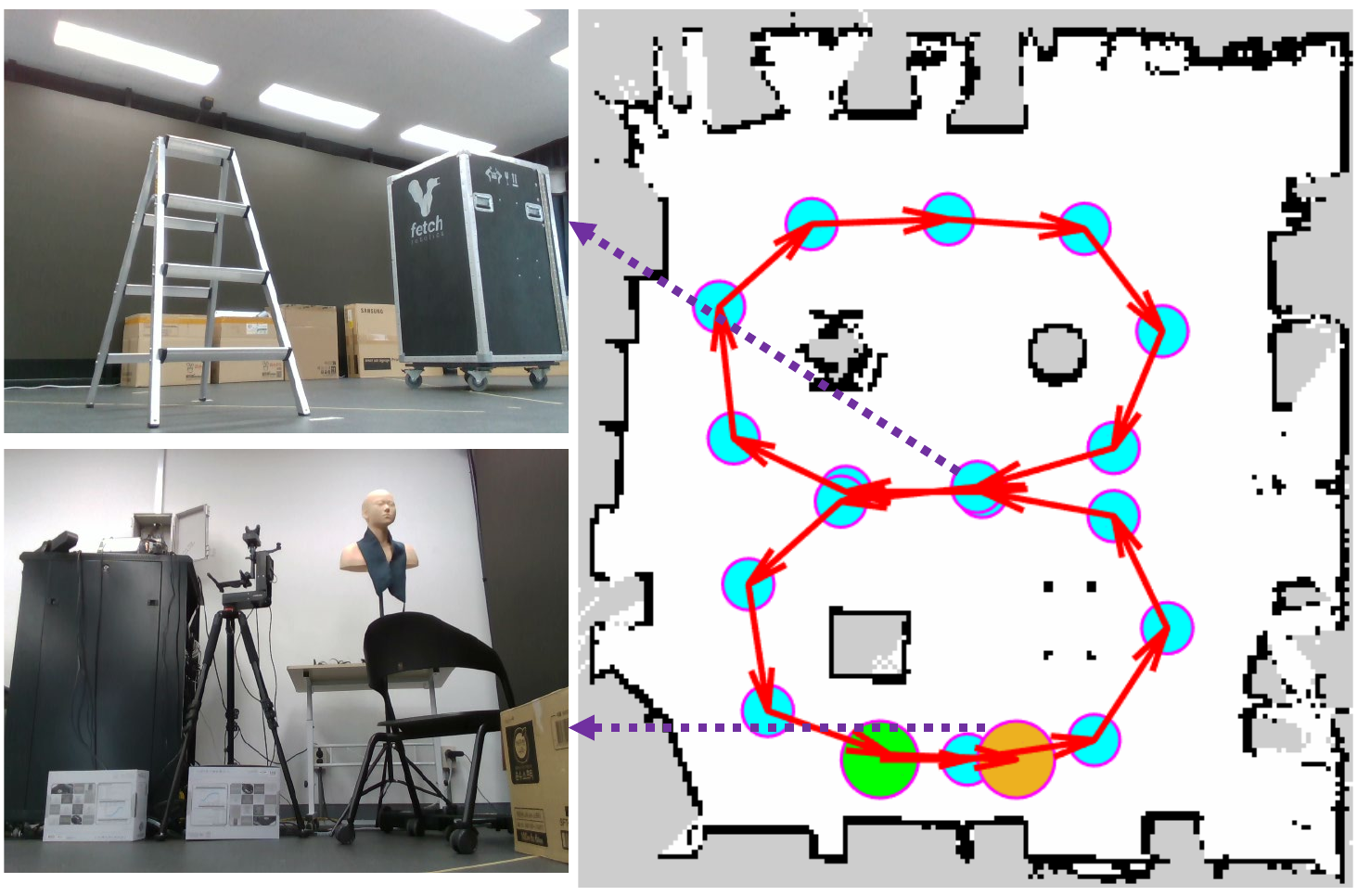}
\caption{T-B9}\label{fig:T-B9}
\end{subfigure}
\\[1.5mm]

\begin{subfigure}[c]{\textwidth}
\centering
\includegraphics[width=0.85\textwidth]{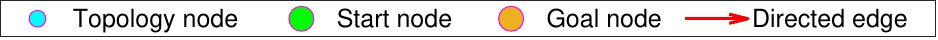}
\end{subfigure}

\caption{\textbf{Topological maps used in the experiments.} From (a) to (i), the path lengths from the start to the goal are 37.1, 54.4, 56.8, 50.4, 65.3, 54.06, 12.14, 18.69, 25.77 in meters, respectively.}
\label{fig:topomaps}
\end{figure*}

\begin{figure*}[!t]
\centering

\begin{subfigure}{0.48\columnwidth}
\centering
\includegraphics[width=\linewidth]{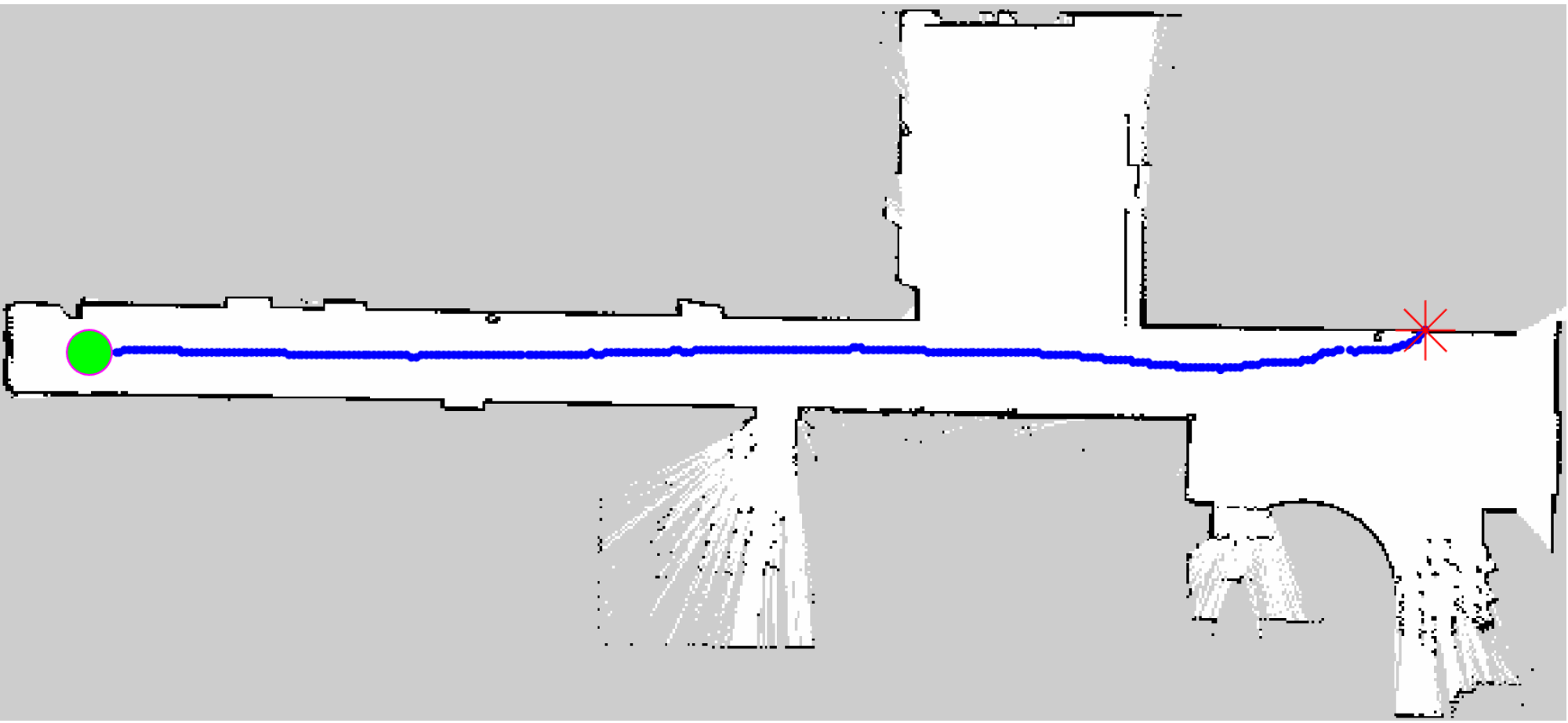}
\caption{NoMaD}
\end{subfigure}
\begin{subfigure}{0.48\columnwidth}
\centering
\includegraphics[width=\linewidth]{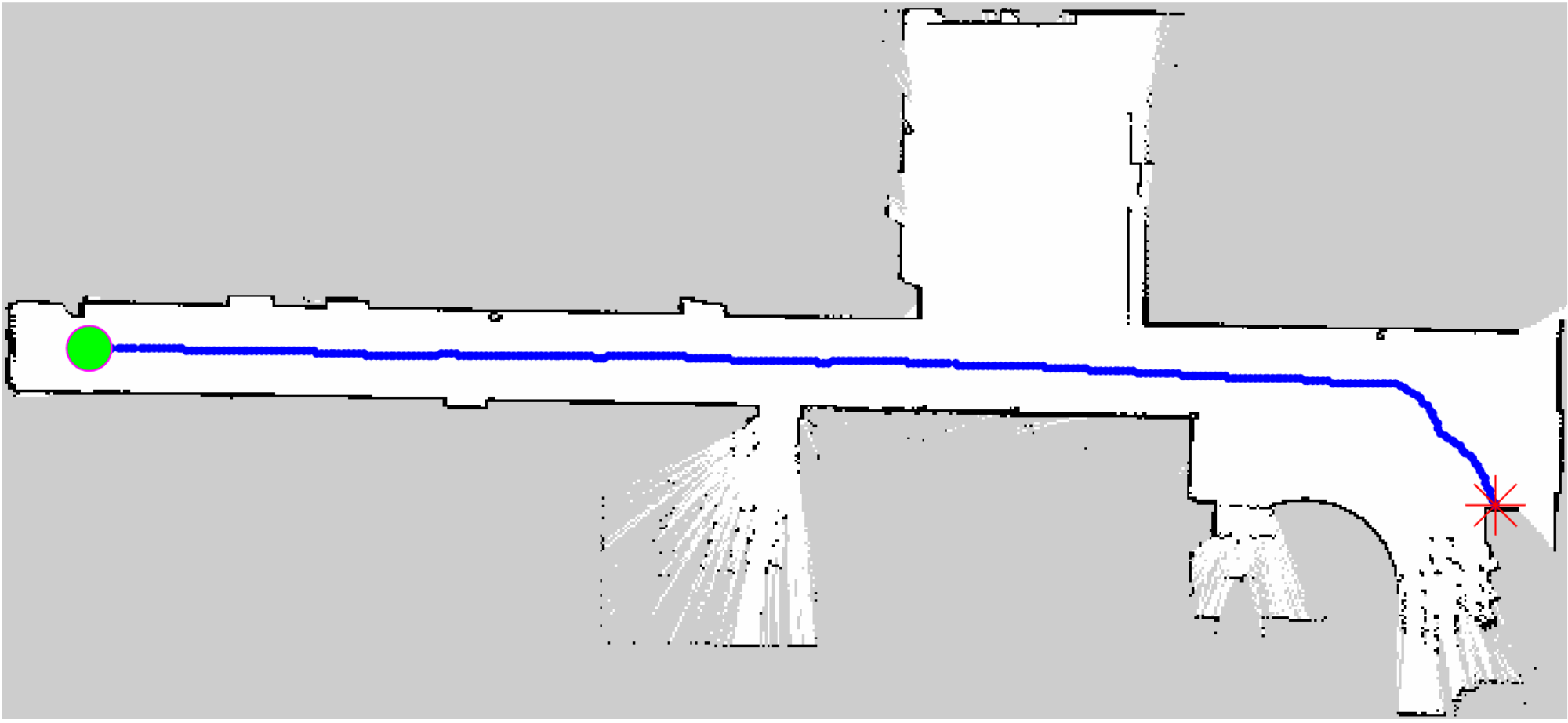}
\caption{ViNT}
\end{subfigure}
\begin{subfigure}{0.48\columnwidth}
\centering
\includegraphics[width=\linewidth]{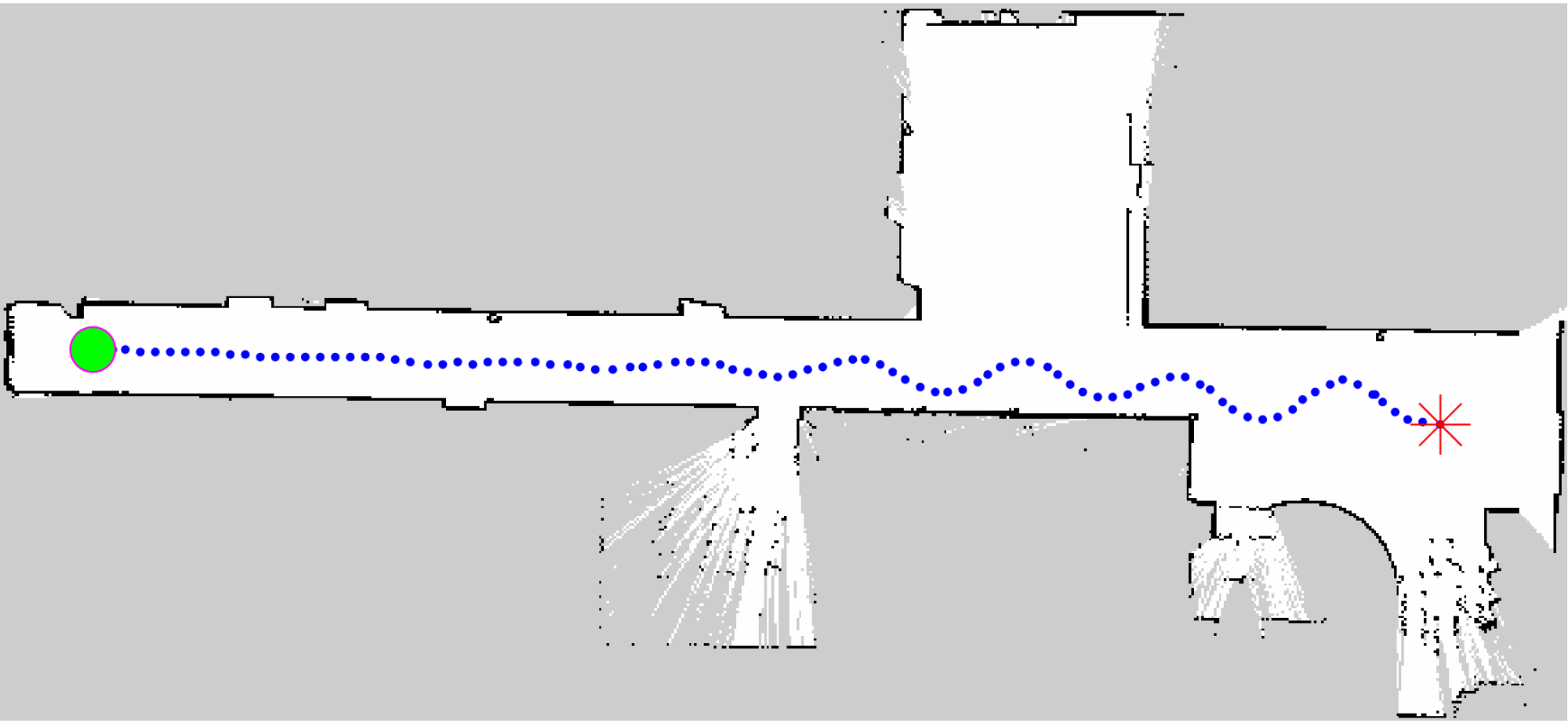}
\caption{NavDP}
\end{subfigure}
\begin{subfigure}{0.48\columnwidth}
\centering
\includegraphics[width=\linewidth]{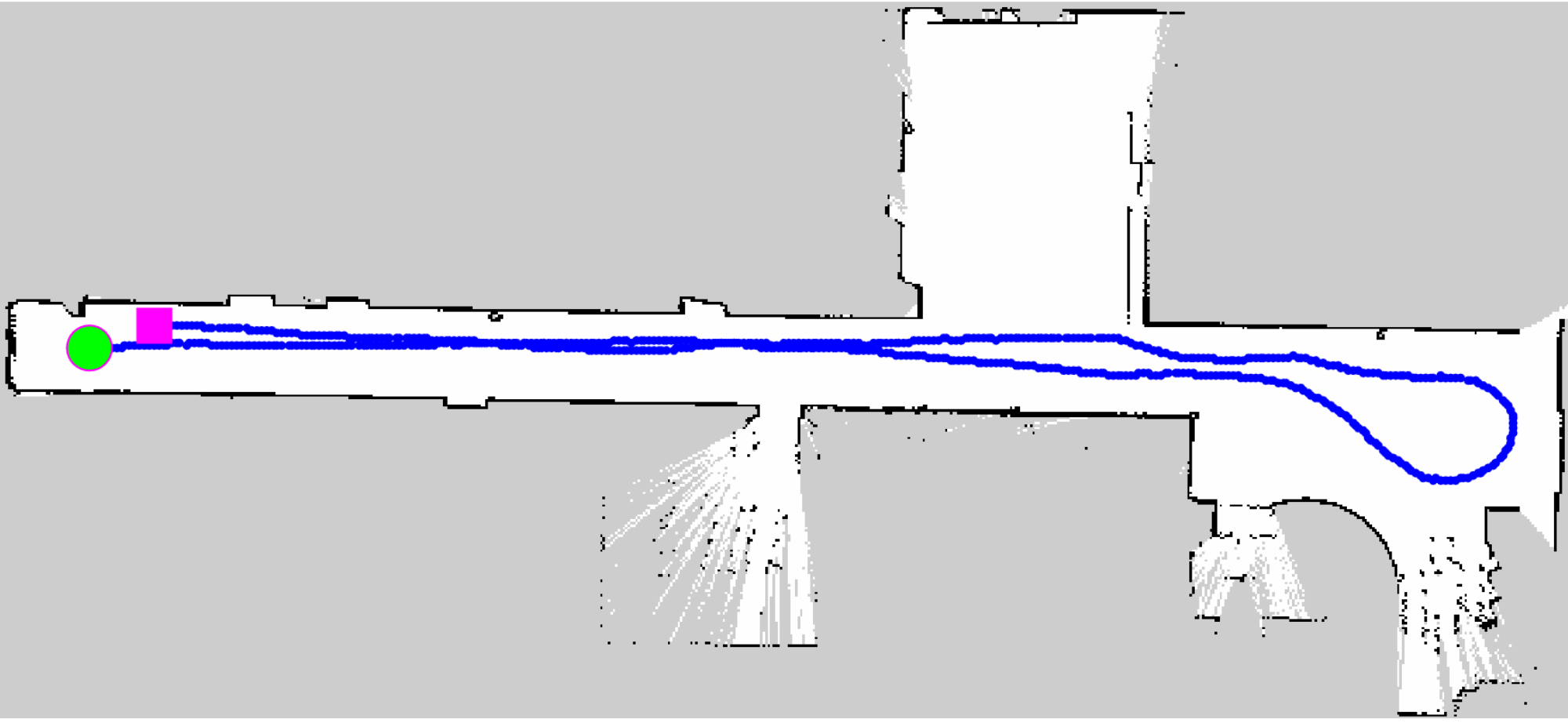}
\caption{DevGRU}
\end{subfigure}


\includegraphics[width=0.6\textwidth]{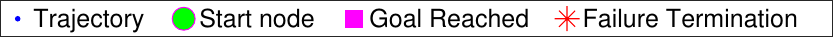}

\caption{\textbf{Qualitative results on the T-B6 scene.} Only our method, DevGRU, successfully completed the full navigation.}
\label{fig:qual_res}

\end{figure*}

\vspace*{-8pt}
\subsection{Navigation Performances against Baselines}
\label{subsec:performance_against_baselines}

We created nine topological maps with diverse {\em topologies} from six locations within the university campus buildings. These places are unseen environments during the training of our network. To evaluate the navigation performance of our method, we compared our method against the three recent state-of-the-art approaches: (1) NoMaD~\cite{SriLev24}, (2) ViNT~\cite{ShaLev23vint}, and (3) NavDP~\cite{CaiPan25}. 
In addition, we created four variants of ViNT and NoMaD by fine-tuning their final checkpoints or retraining them on our 20K collision-free dataset described in Section~\ref{subsec:non-collision_data}. 
These variants are: (4) ViNT-FT-Full (fine-tuning all layers of ViNT), (5) ViNT-FT-FrozEnc (freezing ViNT’s encoder backbone and fine-tuning the remaining layers), (6) ViNT-RandInit (training ViNT from scratch), and (7) NoMaD-FT-Full (fine-tuning all layers of NoMaD).

{ViNT and NoMaD use RGB images, while NavDP requires PG inputs as the navigation target, thereby covering both IG- and PG-based methods.} Table \ref{tab:modality} shows the input and goal modalities of the baseline methods.

On the testing sites, we performed two rounds of topological navigation tasks for each method. Navigation performances were measured in terms of SPL and NC. Fig \ref{fig:topomaps} shows the topological map configuration used for this experiment.

\begin{table*}[ht!]
\centering
\vspace{1pt}
\scriptsize
\setlength{\tabcolsep}{1pt}
\renewcommand{\arraystretch}{1.15}
{
\resizebox{\textwidth}{!}{%
\begin{tabular}{l|cc|cc|cc|cc|cc|cc|cc|cc|cc|cc}
\toprule[1.5pt]
\multirow{2}{*}{Method}
& \multicolumn{2}{c|}{T-B1}
& \multicolumn{2}{c|}{T-B2}
& \multicolumn{2}{c|}{T-B3}
& \multicolumn{2}{c|}{T-B4}
& \multicolumn{2}{c|}{T-B5}
& \multicolumn{2}{c|}{T-B6}
& \multicolumn{2}{c|}{T-B7}
& \multicolumn{2}{c|}{T-B8}
& \multicolumn{2}{c|}{T-B9}
& \multicolumn{2}{c}{Average} \\
\cmidrule(lr){2-3}
\cmidrule(lr){4-5}
\cmidrule(lr){6-7}
\cmidrule(lr){8-9}
\cmidrule(lr){10-11}
\cmidrule(lr){12-13}
\cmidrule(lr){14-15}
\cmidrule(lr){16-17}
\cmidrule(lr){18-19}
\cmidrule(lr){20-21}
& NC $\uparrow$ & SPL $\uparrow$
& NC $\uparrow$ & SPL $\uparrow$
& NC $\uparrow$ & SPL $\uparrow$
& NC $\uparrow$ & SPL $\uparrow$
& NC $\uparrow$ & SPL $\uparrow$
& NC $\uparrow$ & SPL $\uparrow$
& NC $\uparrow$ & SPL $\uparrow$
& NC $\uparrow$ & SPL $\uparrow$
& NC $\uparrow$ & SPL $\uparrow$
& NC $\uparrow$ & SPL $\uparrow$ \\
\midrule

NoMaD
& 0.92 & 0
& 0.19 & 0
& 0.14 & 0
& 0.20 & 0
& 0.07 & 0
& 0.48 & 0
& 0.15 & 0
& 0.34 & 0
& 0.22 & 0
& 0.3 & 0 \\

ViNT
& 0.86 & 0
& 0.48 & 0
& 0.52 & 0
& 0.63 & 0
& 0.05 & 0
& 0.49 & 0
& 0.91 & 0.44
& \bf{1} & \bf{0.92}
& 0.62 & 0
& 0.62 & 0.15 \\

NavDP
& 0.91 & 0.50
& \bf{0.96} & 0.94
& 0.76 & \bf{0.50}
& \bf{1} & \bf{1}
& 0.82 & \bf{1}
& 0.54 & 0
& 0.95 & \bf{1}
& 0.47 & 0.42
& \bf{0.9}4 & 0
& 0.82 & 0.60 \\

ViNT-FT-Full
& \bf{1} & 0.98
& 0.47 & 0
& 0.26 & 0
& 0.25 & 0
& 0.04 & 0
& 0.49 & 0
& 0.49 & 0
& 0.14 & 0
& 0.28 & 0
& 0.38 & 0.11 \\

ViNT-FT-FrozEnc
& \bf{1} & 0.99
& 0.39 & 0
& 0.22 & 0
& 0.24 & 0
& 0.05 & 0
& 0.49 & 0
& 0.15 & 0
& 0.12 & 0
& 0.22 & 0
& 0.32 & 0.11 \\

ViNT-RandInit
& 0.10 & 0
& 0.16 & 0
& 0.18 & 0
& 0.22 & 0
& 0.10 & 0
& 0.08 & 0
& 0.13 & 0
& 0.12 & 0
& 0.22 & 0
& 0.14 & 0 \\

NoMaD-FT-Full
& \bf{1} & \bf{1}
& 0.39 & 0
& 0.22 & 0
& 0.24 & 0
& 0.09 & 0
& 0.49 & 0
& 0.26 & 0
& 0.12 & 0
& 0.22 & 0
& 0.36 & 0.11 \\

DevGRU
& \bf{1} & \bf{1}
& 0.90 & \bf{0.95}
& \bf{0.85} & 0.49
& \bf{1} & 0.99
& \bf{0.87} & 0.98
& \bf{0.83} & \bf{0.97}
& \bf{0.98} & 0.89
& 0.69 & 0.46
& 0.92 & 0.43
& \bf{0.89} & \bf{0.80} \\

\bottomrule[1.5pt]
\end{tabular}%
}
}
\caption{Navigation performance across nine benchmark scenes and the overall average.}
\vspace*{-4pt}
\label{tab:nav_all_scenes}
\end{table*}

\begin{table}[htb!]
\centering
\small
\begin{tabular}{l|cc}
\toprule[1.5pt]
\makecell{\textbf{Method}} &
\makecell{\textbf{Input modality} } &
\makecell{\textbf{Goal modality} } \\
\midrule
NoMaD        & RGB & RGB \\
ViNT         & RGB & RGB \\
NavDP        & RGB + Depth & Point Goal (PG) \\
{DevGRU} & Depth + Egomotion & Depth + PG \\
\bottomrule[1.5pt]
\end{tabular}
\caption{\textbf{\textbf{Input and goal modalities of each method}} }
\vspace*{-8pt}
\label{tab:modality}
\end{table}

{Table \ref{tab:nav_all_scenes} reports detailed per-scene results across the nine scenes and their averages.
Overall, our method outperforms all baseline approaches.}
We also present a qualitative example in Figure \ref{fig:qual_res}, which illustrates the navigation results on the scene shown in {Figure \ref{fig:T-B6}}. {In summary, our approach significantly outperforms ViNT and NoMaD, and also achieves improved performance over NavDP. Despite this,} our model is substantially more compact than NavDP in terms of parameter count and inference time, as discussed in the following section.

\subsection{Model Efficiency}
\label{subsec:model_size_analysis}

Aside from the superior navigation performance of our method, it is also important to highlight that our model requires a relatively small number of {trainable} parameters. A lightweight model is particularly important for deployment in mobile robot navigation, where computational and memory resources are limited. To evaluate model compactness, we perform a parameter count analysis for each method. Table~\ref{tab:param_size} reports the total number of trainable parameters and the mean inference time of the compared models.


The total number of parameters required by our method, obtained by summing the action predictor (14.77M) and the collision predictor (4.66M), is substantially smaller than that of NavDP. Overall, our model is approximately 53\% and 700\% more parameter-efficient than ViNT and  NavDP, respectively. This compactness is achieved mainly by using depth instead of RGB (\ie reduced dimensionality) and a lightweight recurrent policy head based on a GRU, rather than the transformer model used by the competitive methods.

\begin{table}[htb!]
\centering
\setlength{\tabcolsep}{8pt}
\renewcommand{\arraystretch}{1.1}
\begin{tabular}{c|cc}
\toprule[1.5pt]
Model & Parameters (M) $\downarrow$ & Mean Inf. Time (ms) $\downarrow$ \\
\midrule
NoMaD & \textbf{19.05} \if23.52\fi  & 123.46 \\
ViNT  & 29.67 \if23.36\fi  & 176.57 \\
NavDP & 135.72 & 860.18 \\
{DevGRU}  & 19.44 & \textbf{50.01} \\
\bottomrule[1.5pt]
\end{tabular}
\caption{\textbf{Model efficiency comparison in terms of parameter size and inference time.}} 
\vspace*{-12pt}
\label{tab:param_size}
\end{table}




\vspace{-10pt}
\subsection{Ablation Study}
\label{subsec:ablation_study}
\begin{figure}[htb]
\centering
\includegraphics[width=1\columnwidth]{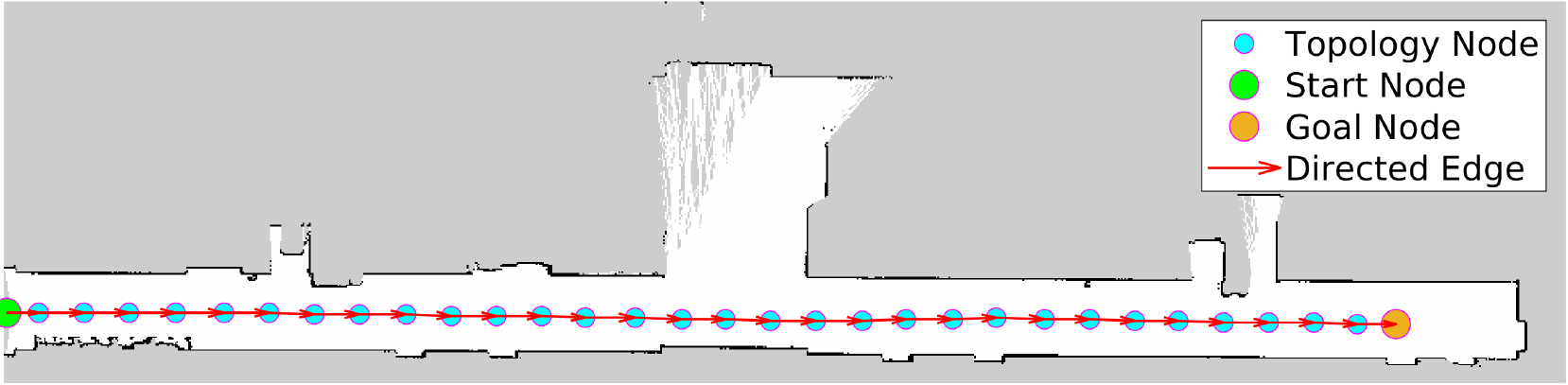}
\includegraphics[width=1\columnwidth]{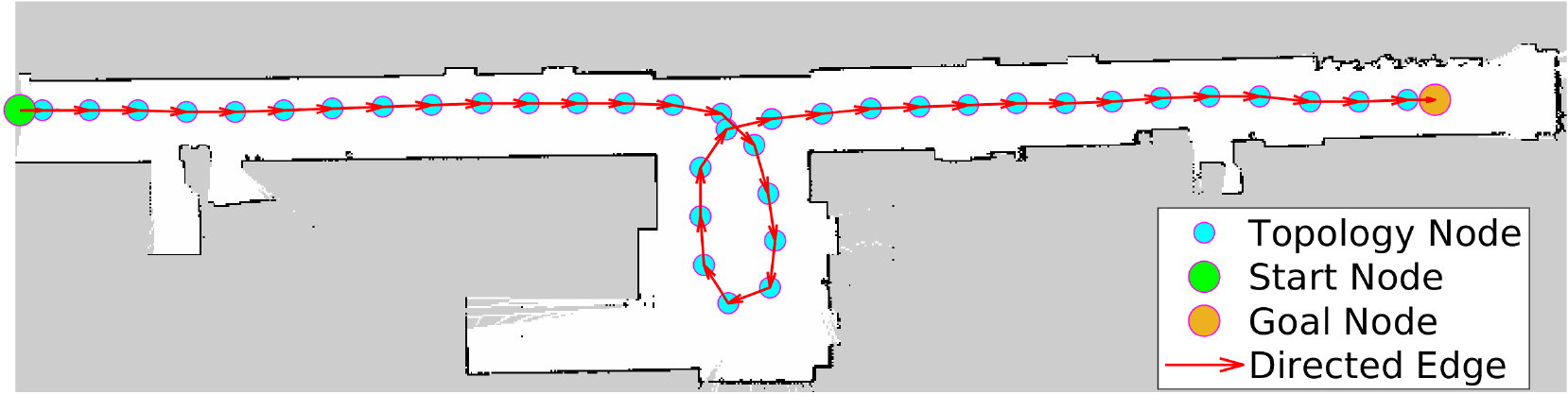}
\caption{\textbf{The two topological maps used for the ablation study:} (Top) T-A1 and (bottom) T-A2. The topological maps are shown together with occupancy maps for visualization and quantitative analysis.}
\vspace{-10pt}
\label{fig:ablation_topomaps}
\end{figure}

Our method contains two critical features that distinguish it from existing approaches, including GNM~\cite{ShaLev23gnm}, ViNT~\cite{ShaLev23vint}, and NoMaD~\cite{SriLev24}. These features are (1) collision-aware waypoint prediction and (2) an SG pose correction mechanism that facilitates long-distance navigation. To highlight the importance of these components, we conducted an ablation study. First, we trained a \textit{{DevGRU-Base}} model on the collision-free dataset alone, without SG-pose correction. Next, we constructed a \textit{{DevGRU-No-SG-Correction}} model trained on both collision and collision-free datasets, without SG-pose correction. Finally, we evaluated the \textit{{DevGRU-Full}} model, which incorporates all components of the proposed system. To evaluate these models, we constructed two extra topological maps, as shown in Fig.~\ref{fig:ablation_topomaps}.

\begin{table}[htb!]
\centering
\setlength{\tabcolsep}{6pt}
\renewcommand{\arraystretch}{1.1}

\begin{tabular}{l|cc}
\toprule[1.5pt]
\makecell{Model} 
& \makecell{{NC}  $\uparrow$} 
& \makecell{SPL $\uparrow$} \\
\midrule
DevGRU-Base             & 0.82 & 0.25 \\
DevGRU-No-SG-Correction       & 0.73 & 0.49 \\
DevGRU-Full            & \textbf{0.99} & \textbf{0.73}\\
\bottomrule[1.5pt]
\end{tabular}
\caption{ \textbf{Ablation study results.} Higher is better for Node Coverage (NC) and SPL.}
\label{tab:ablation}
\end{table}

\begin{figure}[htb!]
\centering
\includegraphics[width=1\columnwidth]{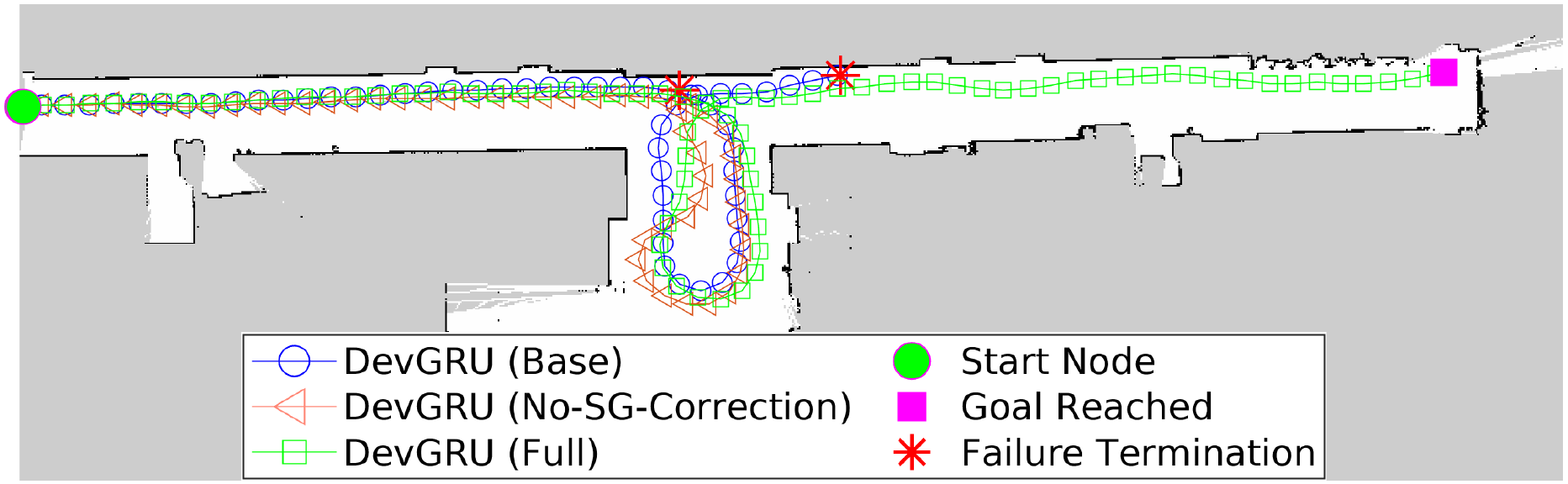}

\caption{\textbf{Navigation results of the three model variants on T-A2.} Only \textit{DevGRU-Full} successfully completed the trajectory.}
\vspace{-10pt}
\label{fig:ablation_res}
\end{figure}

\begin{figure}[htb!]
\centering
\includegraphics[width=\columnwidth]{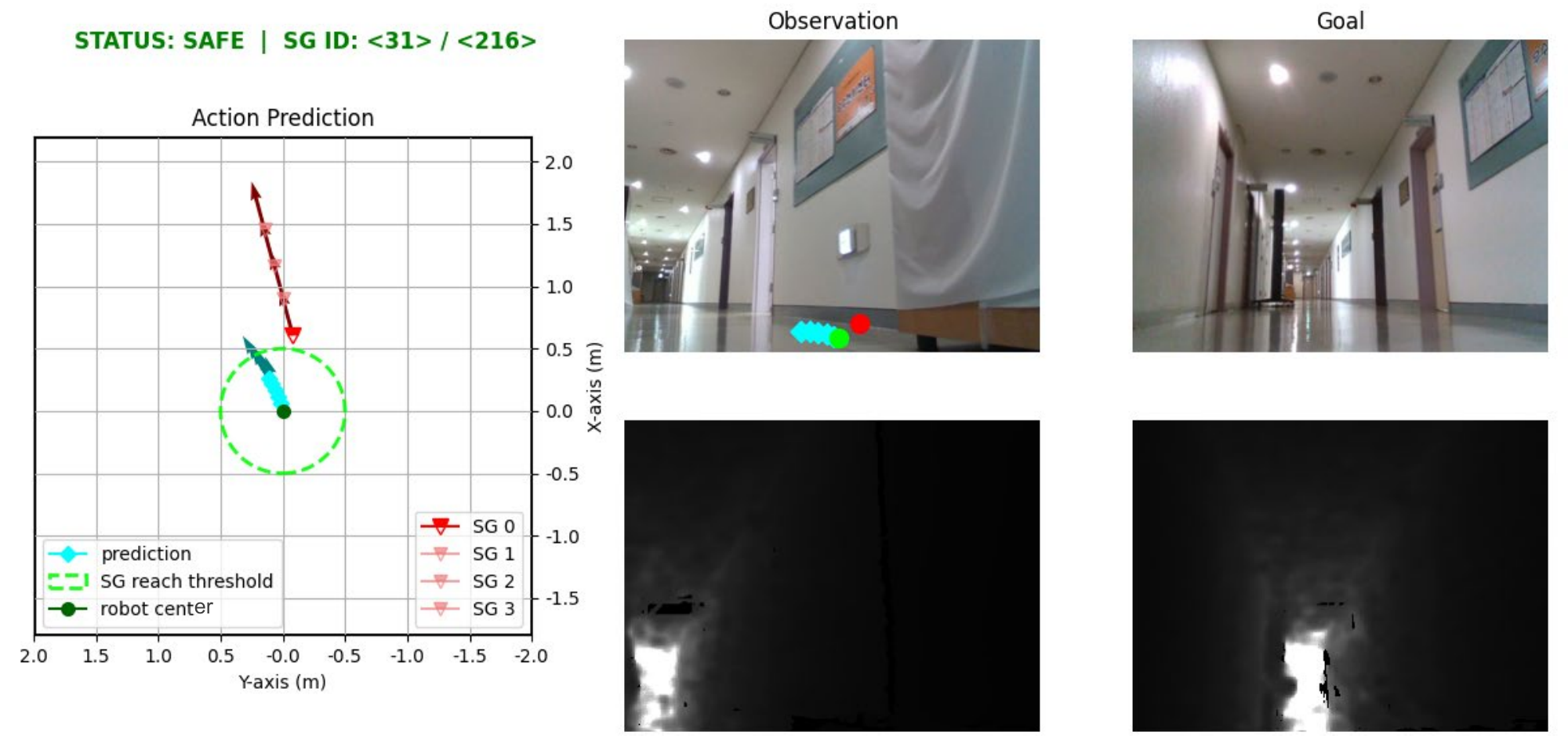}

\caption{\textbf{The collision avoidance instance using the full DevGRU.} The leftmost column shows the future SGs $\{\tilde{v}_i,\ldots,\tilde{v}_{i+3} \}$ in red and the collision-aware waypoints $\hat{\mathcal{W}_t}$ in cyan w.r.t. the robot pose $\mathbf{p}_t$. The center column shows an RGB–depth image pair overlaid with $\hat{\mathcal{W}_t}$ and $\tilde{v}_i$. The rightmost column shows the RGB–depth image pair corresponding to $v_i$.}
\label{fig:coll_avoidance}
\end{figure}

Fig.~\ref{fig:ablation_res} shows navigation instances of the three model variants in one of the test environments {shown in Figure \ref{fig:ablation_topomaps}}. Only the \textit{DevGRU-Full} model successfully reached the final goal, while the other variants did not. Figure \ref{fig:coll_avoidance} shows an obstacle avoidance example guided by our \textit{DevGRU-Full} model. In this example, the robot successfully generates collision-free waypoints to avoid the wall obstacle ahead.

Table \ref{tab:ablation} reports the quantitative results averaged over all navigation experiments, highlighting the importance of the collision avoidance and SG-pose correction mechanisms. 
It is observed that the collision avoidance feature alone allows the robot to avoid only immediate obstacles. Without the SG-pose correction mechanism, the robot repeatedly aims at the SG, accumulating pose error and eventually colliding with obstacles.

\section{Conclusions}
\label{Conclusions}
In this paper, we introduce a depth-based, visual navigation method, DevGRU. DevGRU consists of an action predictor and a collision predictor that actively use depth images to generate a collision-aware future trajectory. Unlike existing state-of-the-art models, our approach can predict the SG pose, enabling correction of accumulated SG pose errors during navigation. This capability enables the robot to navigate over longer horizons than baseline approaches. Furthermore, the proposed model adopts a lightweight architecture with a relatively small number of parameters, making it suitable for deployment on resource-constrained embedded mobile robots.



\bibliographystyle{IEEEtran}
\bibliography{reference.bib}

\newpage

\vfill

\end{document}